\documentclass[runningheads]{llncs}

\usepackage{accv}

\usepackage{accvabbrv}

\usepackage{graphicx}
\usepackage{booktabs}
\usepackage{amsmath,amssymb}
\usepackage{multirow}
\usepackage{makecell}
\usepackage{subcaption}
\usepackage{array}
\usepackage{colortbl}  % xcolor (with dvipsnames) is already loaded by accv.sty; colortbl adds \rowcolor
\usepackage{float}
\newcolumntype{L}[1]{>{\centering\arraybackslash}p{#1}}
\newcolumntype{M}[1]{>{\centering\arraybackslash}m{#1}}
\newcolumntype{C}[1]{>{\centering\arraybackslash}m{#1}}
\usepackage[ruled,linesnumbered]{algorithm2e}
\definecolor{oursrow}{rgb}{0.90,0.94,1.0}
\definecolor{bestgreen}{RGB}{220,240,220}

\usepackage[accsupp]{axessibility}  % Improves PDF readability for those with disabilities.

\usepackage[breaklinks,colorlinks,citecolor=accvblue]{hyperref}

\begin{document}

% ---------------------------------------------------------------
\title{AirflowAttack: Thermal-Airflow Adversarial Perturbations against Infrared Remote-Sensing Vision-Language Models}

\titlerunning{Thermal-Airflow Attack against IR Remote Sensing VLMs}

\author{Cong Su\inst{2} \and Jiaju Han\inst{1} \and Xuemeng Sun\inst{1} \and Chengyin Hu\inst{1} \and Qike Zhang\inst{1} \and Jiujiang Guo\inst{2} \and Yiwei Wei\inst{1} \and Jiahuan Long\inst{3}}
\authorrunning{C. Su et al.}
\institute{China University of Petroleum-Beijing at Karamay, Karamay, Xinjiang, China
\and Tianjin University, Tianjin, China
\and Shanghai Jiao Tong University, Shanghai, China}

\maketitle

\begin{abstract}
Vision-language models (VLMs) are increasingly deployed on infrared (IR) remote sensing imagery in security-critical settings, yet their adversarial robustness remains unexamined. We present AirflowAttack, to our knowledge the first adversarial attack for IR remote-sensing VLMs and the first to weaponize thermal-airflow turbulence as the perturbation prior. A lightweight generator synthesizes a single input-agnostic perturbation regularized toward physically plausible airflow patterns. Optimized on one surrogate CLIP model, it attains a mean zero-shot scene-classification attack success rate (ASR, the fraction of samples whose top-1 class changes) of 48.5\% across five diverse CLIP backbones, far exceeding four IR-specific physical baselines (27.7--37.0\%). Applied to six state-of-the-art VLMs, it cuts scene-classification accuracy by up to 38.2\% (relative)---yet paradoxically makes some models \emph{more} confident in their IR analysis, confabulating the perturbation as genuine thermal evidence such as temperature gradients and convection. Ablations show the airflow prior raises physical plausibility at no measurable cost to attack success. Together with a benchmark spanning eleven models and four tasks, these findings expose critical vulnerabilities in the rapidly expanding IR VLM ecosystem.
\end{abstract}

\begin{figure}[t]
  \centering
  \includegraphics[width=\textwidth]{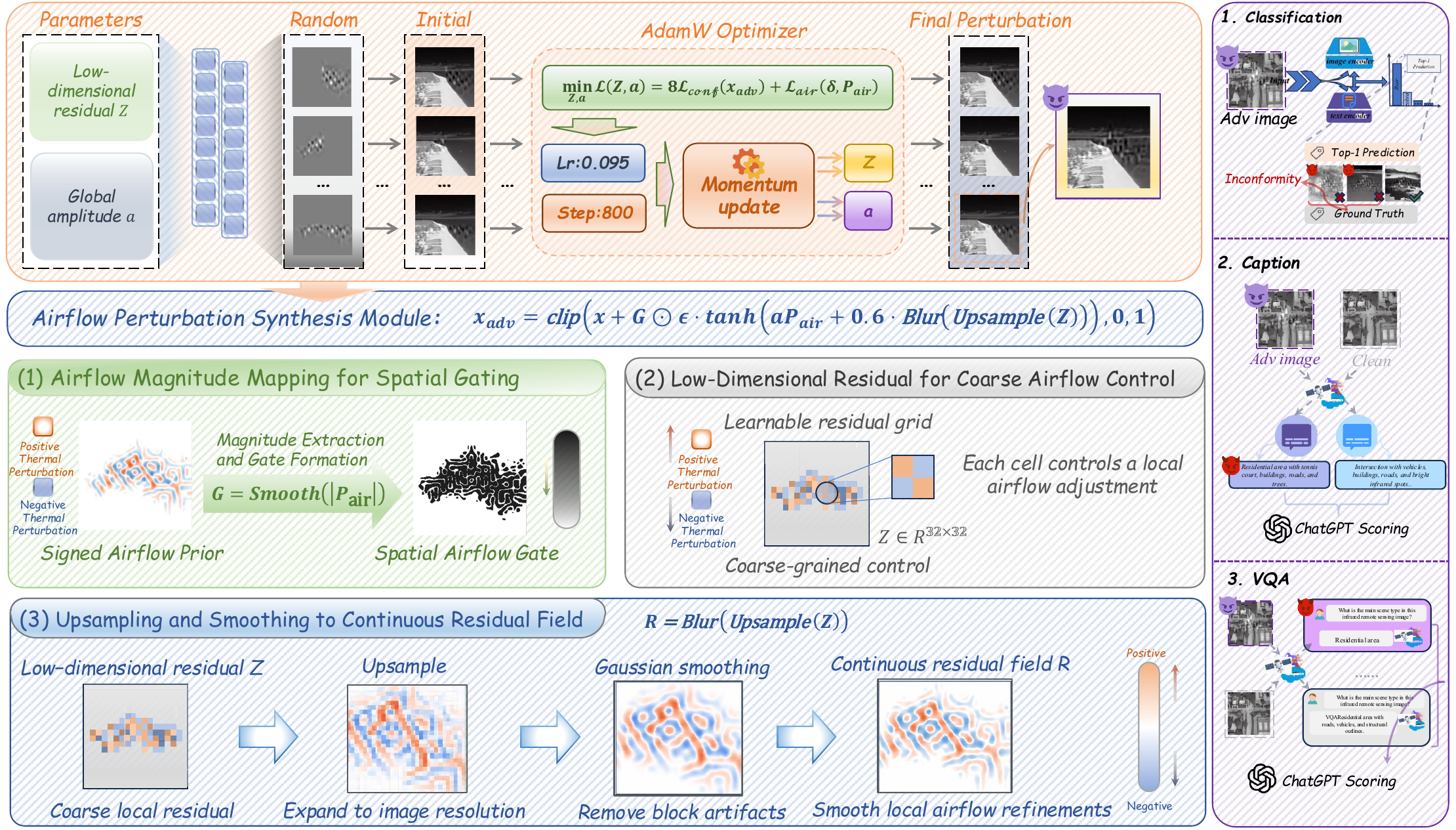}
  \caption{Overview of AirflowAttack. A lightweight generator $\mathcal{G}_\theta$ maps a low-dimensional latent code to a single-channel thermal-airflow perturbation, optimized on a surrogate IR-finetuned CLIP model under an $L_\infty \leq \varepsilon$ constraint using a confidence loss $\mathcal{L}_{\text{conf}}$ and an airflow-correlation loss $\mathcal{L}_{\text{air}}$. The resulting perturbation transfers, without target-model access, to five CLIP backbones and six VLMs across four vision-language tasks.}
  \label{fig:framework}
\end{figure}

\section{Introduction}
\label{sec:intro}

Infrared remote sensing underpins critical applications from disaster monitoring and environmental surveillance to military reconnaissance, operating under conditions where visible-spectrum imaging fails---nighttime, fog, smoke, and thermal camouflage detection. The recent adaptation of vision-language models (VLMs) to the IR domain promises a step change in automated scene understanding: models such as GeoRSCLIP~\cite{georsclip}, RemoteCLIP~\cite{remoteclip}, and RS5M~\cite{rs5m} can now jointly reason about IR imagery and natural language, enabling open-vocabulary retrieval, descriptive captioning, and visual question answering over thermal scenes. However, the security implications of deploying VLMs in IR-sensitive contexts remain entirely unexamined.

Adversarial attacks---imperceptible input perturbations that cause models to fail---have been extensively studied in the RGB domain, spanning white-box~\cite{goodfellow2015}, black-box~\cite{papernot2017}, and universal~\cite{moosavi2017} regimes. Yet IR imagery differs fundamentally from RGB: thermal sensors capture emitted radiation rather than reflected light, producing single-channel intensity maps governed by Planck's law where pixel values encode physical temperature. This physical grounding both constrains and motivates a new class of attacks: rather than crafting arbitrary digital noise, an adversary can simulate physically plausible thermal phenomena---such as airflow-induced temperature distortions---that are simultaneously harder to detect and more likely to transfer across models.

In this paper, we introduce AirflowAttack, to the best of our knowledge the first adversarial attack designed for IR remote-sensing VLMs. While universal perturbations have been studied for unimodal remote-sensing classifiers~\cite{xu2023uaprs} and for RGB vision-language models~\cite{advclip,coattack}, and while simulated atmospheric phenomena such as haze~\cite{gao2021advhaze} and weather~\cite{schmalfuss2023downpour} have been repurposed as attacks in the RGB domain, none of these targets the thermal-IR modality, the remote-sensing VLM setting, or thermal-airflow turbulence as the perturbation prior---the specific intersection our work addresses. Our approach synthesizes an input-agnostic adversarial perturbation by modeling temperature fluctuations induced by thermal airflow turbulence. Unlike conventional pixel-space UAPs that produce unstructured noise, the proposed perturbation mimics the spatially correlated, physically grounded patterns of atmospheric thermal mixing---making it simultaneously effective, transferable across architectures, and physically interpretable as a natural thermal phenomenon (see \cref{fig:framework}).

We conduct a comprehensive adversarial robustness evaluation spanning five CLIP-family backbones (OpenAI-CLIP-B32, OpenAI-CLIP-L14, OpenCLIP-B32, RemoteCLIP-B32, GeoRSCLIP-B32) and six state-of-the-art VLMs (Qwen2.5-VL-7B, InstructBLIP, LLaVA-1.5, LLaVA-1.6, GeoChat, H2RSVLM) across four vision-language tasks on a dedicated 1,000-sample infrared test set. Our contributions are:

\begin{itemize}
\item We propose AirflowAttack, to our knowledge the first adversarial attack for IR remote-sensing VLMs, synthesizing a transferable thermal-airflow perturbation via a lightweight generator.
\item Across five CLIP backbones and six VLMs, a single surrogate-optimized perturbation attains 48.5\% mean ASR, exceeding four IR-specific baselines and transferring without target access.
\item Extensive ablations reveal the mechanisms of airflow-based transfer, including a paradoxical rise in some models' IR-cue confidence under attack.
\end{itemize}

\section{Related Work}
\label{sec:related}

\subsection{Infrared Remote Sensing Vision-Language Models}
CLIP~\cite{radford2021} showed that contrastive image-text pretraining yields strong zero-shot transfer, a paradigm extended to remote sensing by RemoteCLIP~\cite{remoteclip}, GeoRSCLIP~\cite{georsclip}, and RS5M~\cite{rs5m}, and by general-purpose VLMs such as LLaVA-1.5/1.6~\cite{liu2023llava,liu2024llava}, InstructBLIP~\cite{instructblip}, Qwen2.5-VL~\cite{qwen25vl}, GeoChat~\cite{geochat}, and H2RSVLM~\cite{h2rsvlm} applied to RS scene classification, detection, VQA, and captioning. These models are developed and evaluated under benign conditions. While the robustness of unimodal RS classifiers has been studied~\cite{xu2023uaprs,xu2023aisecurity}, and very recent work crafts transferable adversarial examples for RS object recognition~\cite{liu2025rstransfer} and attacks infrared VLMs with physical patches~\cite{hu2026irvlmpatch}, the robustness of \emph{IR} remote-sensing VLMs to input-agnostic perturbations remains unexamined.

\subsection{Adversarial Attacks}
Since neural networks were shown vulnerable to imperceptible perturbations~\cite{szegedy2014}, white-box attacks (FGSM~\cite{goodfellow2015}, C\&W~\cite{carlini2017}, PGD~\cite{madry2018}, AutoAttack~\cite{croce2020}) and transfer-boosting methods (MI-FGSM~\cite{dong2018}, DIM~\cite{xie2019dim}) have been widely studied. Universal perturbations~\cite{moosavi2017} fool a model across inputs with a single pattern, and have been extended to remote sensing~\cite{xu2023uaprs} and to CLIP/VLP models (AdvCLIP~\cite{advclip}, Co-Attack~\cite{coattack}, and recent targeted VLM attacks~\cite{cao2025targeted}). These target unimodal RGB classifiers or RGB CLIP; physically realizable RGB attacks~\cite{brown2017,athalye2018} likewise exploit color/texture cues absent in thermal imagery. Our perturbation instead targets the single-channel thermal modality and is parameterized to resemble a physical thermal phenomenon.

\subsection{Physical Attacks on Thermal Imaging}
A parallel line of work explores physical attacks specific to thermal sensors. Methods include projecting thermal patterns using controlled heat sources~\cite{thermalpatch}, physically adversarial infrared patches with learnable shapes and locations~\cite{zhu2022irpatch}, adversarial infrared curves and grid patterns against pedestrian detectors~\cite{hu2024adversarial,tiliwalidi2025advgrid}, wearable hot/cold blocks that fool thermal detectors~\cite{hotcold}, exploiting sensor non-uniformity~\cite{sensordrift}, and introducing fixed-pattern noise~\cite{stripingnoise}. A recent survey~\cite{wei2023irsurvey} provides a broader taxonomy of IR-specific adversarial methods. A closely related line simulates a physical atmospheric phenomenon and repurposes it as an adversarial perturbation---adversarial haze~\cite{gao2021advhaze}, adversarial weather~\cite{schmalfuss2023downpour}, and, concurrently, physically-induced atmospheric perturbations for RS classification~\cite{zhuang2026atmospheric}; our airflow-turbulence perturbation is a new instance of this template, uniquely targeting the thermal-IR VLM setting. Atmospheric turbulence, caused by spatial and temporal variations in air refractive index due to temperature gradients, is a well-known degradation in long-range thermal imaging~\cite{turbulence}; learned turbulence simulation models~\cite{mao2021turbulence} have improved the fidelity of synthetic turbulence. However, prior work treats turbulence as a nuisance to be corrected, not as an adversarial primitive to be exploited. Our work is the first to synthesize thermal airflow turbulence patterns as a deliberate universal perturbation optimized for attacking multimodal IR models.

\subsection{VLM Robustness}
Recent studies have begun examining the adversarial robustness of VLMs~\cite{schlarmann2023,zhao2023evaluating}. Adversarial images can cause VLMs to produce hallucinated captions~\cite{vlmattackcaption}, and visual adversarial perturbations can transfer to the language modality~\cite{vlmtransfer}. Set-of-mark prompting~\cite{zhang2024som} further reveals how VLMs attend to image regions, which is directly relevant to understanding why spatially localized perturbations matter. However, existing work focuses almost exclusively on RGB inputs and generic multimodal reasoning. The IR modality introduces distinct challenges: single-channel intensity encoding, different feature statistics, and physical constraints on plausible perturbations. To the best of our knowledge, this work is the first to study adversarial robustness for IR remote-sensing VLMs.

\section{Method}
\label{sec:method}

We propose AirflowAttack, a framework for generating universal adversarial perturbations that simulate thermal airflow turbulence to attack IR vision-language models. The key insight is that physically interpretable thermal patterns---unlike arbitrary pixel noise---exploit domain-specific feature representations and transfer more effectively across architectures.

\subsection{Threat Model}
\label{sec:threat}

We consider a gray-box adversary with access to a surrogate CLIP model (OpenAI-CLIP-B32, IR-finetuned) but no access to target models, their parameters, or training data. The adversary can perturb input IR images before they are processed by downstream VLMs. The perturbation must satisfy an $L_\infty$ constraint: $\|\delta\|_\infty \leq \varepsilon$, where $\varepsilon = 100$ in pixel intensity space (out of 255). The adversary's goal is to craft a single universal perturbation $\delta$ that, when added to any IR image $x$, causes incorrect scene classification, captioning, or VQA outputs across multiple target models. This is a realistic threat: IR sensors deployed in the field may process adversarially perturbed inputs before human or automated analysis.

\subsection{Thermal Airflow Perturbation Model}
\label{sec:perturbation_model}

Rather than optimizing $\delta$ directly in pixel space, we parameterize the perturbation through a lightweight generative model $\mathcal{G}_\theta(z)$ that maps a low-dimensional latent vector $z \in \mathbb{R}^d$ to a full-resolution perturbation pattern. This parameterization serves three purposes: it regularizes the perturbation toward spatially coherent patterns, it reduces the optimization search space, and it enables the integration of physical priors through architectural design.
The generator consists of a series of transposed convolution layers with residual connections, mapping a latent code of dimension $d = 32$ to a single-channel output matching the input resolution. We employ a residual ratio of $r = 0.60$, meaning that 60\% of the perturbation energy passes through the residual pathway, which we find empirically balances attack strength against physical plausibility (see \cref{sec:ablations}).

For each training iteration, a latent vector $z$ is sampled and decoded to produce the perturbation $\delta = \mathcal{G}_\theta(z)$. The perturbation is then scaled to satisfy the $L_\infty$ constraint: $\delta \leftarrow \text{clip}(\delta, -\varepsilon, \varepsilon)$. The adversarial image is formed as $x^{\text{adv}} = x + \delta$.

\subsection{Optimization Objective}
\label{sec:objective}

The optimization objective combines an attack loss that drives retrieval failure with a correlation loss that enforces physical plausibility. Given an IR image $x$, its paired caption $t^+$, and a set of distractor captions $\mathcal{T}^-$, we define the confidence (contrastive image--text alignment) loss:

\begin{equation}
\mathcal{L}_{\text{conf}} = -\log \frac{\exp(\tau \cdot \text{sim}(f_I(x^{\text{adv}}), f_T(t^+)))}{\sum_{t \in \{t^+\} \cup \mathcal{T}^-} \exp(\tau \cdot \text{sim}(f_I(x^{\text{adv}}), f_T(t)))}
\label{eq:conf_loss}
\end{equation}

where $f_I$ and $f_T$ are the image and text encoders of the surrogate CLIP model, $\text{sim}(\cdot,\cdot)$ denotes cosine similarity, and $\tau$ is a temperature parameter. Minimizing $\mathcal{L}_{\text{conf}}$ reduces the similarity between the adversarial image embedding and its correct caption while increasing similarity with distractors.

To encourage the perturbation to resemble natural thermal airflow, we introduce an airflow correlation loss:

\begin{equation}
\mathcal{L}_{\text{air}} = 1 - \text{Corr}(\delta, \mathcal{A})
\label{eq:air_loss}
\end{equation}

where $\mathcal{A}$ is a bank of synthetic thermal airflow templates generated by simulating atmospheric turbulence via randomized heat kernel convolution, and $\text{Corr}(\cdot,\cdot)$ denotes spatial Pearson correlation. This term penalizes perturbations that deviate from physically plausible airflow patterns.

The total loss is a weighted combination:

\begin{equation}
\mathcal{L} = \alpha \cdot \mathcal{L}_{\text{conf}} + \beta \cdot \mathcal{L}_{\text{air}}
\label{eq:total_loss}
\end{equation}

with $\alpha = 8$ and $\beta = 2$, determined through ablation (\cref{sec:ablations}). Optimization proceeds via projected gradient descent over the generator parameters $\theta$ and latent code $z$ for 800 steps with learning rate $\eta = 0.055$. \Cref{alg:airflow} summarizes the full procedure.

\begin{algorithm}[t]
\SetAlgoLined
\DontPrintSemicolon
\SetCommentSty{textnormal}
\KwIn{Surrogate CLIP encoders $f_I, f_T$; IR pairs $\{(x,t^+)\}$ with distractors $\mathcal{T}^-$; airflow templates $\mathcal{A}$; budget $\varepsilon$, steps $T$, rate $\eta$, weights $\alpha,\beta$}
\KwOut{Universal perturbation $\delta^\star$}
$\theta, z \leftarrow \textsc{Init}$\tcp*{generator params and latent code}
\For{$i = 1 \ldots T$}{
  $\delta \leftarrow \mathrm{clip}\big(\mathcal{G}_\theta(z),\, -\varepsilon,\, \varepsilon\big)$\tcp*{decode, project to $L_\infty$ ball}
  $x^{\text{adv}} \leftarrow x + \delta$\;
  $\mathcal{L}_{\text{conf}} \leftarrow -\log \dfrac{\exp(\tau\,\mathrm{sim}(f_I(x^{\text{adv}}), f_T(t^+)))}{\sum_{t \in \{t^+\}\cup\mathcal{T}^-}\exp(\tau\,\mathrm{sim}(f_I(x^{\text{adv}}), f_T(t)))}$\tcp*{attack loss}
  $\mathcal{L}_{\text{air}} \leftarrow 1 - \mathrm{Corr}(\delta, \mathcal{A})$\tcp*{airflow-plausibility loss}
  $\mathcal{L} \leftarrow \alpha\,\mathcal{L}_{\text{conf}} + \beta\,\mathcal{L}_{\text{air}}$\;
  $(\theta, z) \leftarrow (\theta, z) - \eta\,\nabla_{(\theta,z)}\mathcal{L}$\tcp*{gradient step}
}
$\delta^\star \leftarrow \mathrm{clip}\big(\mathcal{G}_\theta(z),\, -\varepsilon,\, \varepsilon\big)$\;
\Return{$\delta^\star$}\;
\caption{AirflowAttack universal perturbation optimization}
\label{alg:airflow}
\end{algorithm}

\subsection{Surrogate-to-Target Transfer}
\label{sec:transfer}

Once optimized on the surrogate model, the final UAP $\delta^* = \mathcal{G}_{\theta^*}(z^*)$ is applied directly to any target IR image. No access to target models is required at deployment time. The perturbation transfers effectively because: (i) the airflow pattern encodes domain-general thermal features that multiple IR-trained models have learned to associate with atmospheric conditions; (ii) the universal nature of the perturbation means it does not overfit to surrogate-specific decision boundaries; and (iii) the $L_\infty$ constraint ensures the perturbation magnitude is consistent across inputs, avoiding the brittleness of per-image attacks.

\section{Experiments}
\label{sec:experiments}

\subsection{Experimental Setup}
\label{sec:setup}

\textbf{Models.} We evaluate five CLIP-family backbones spanning different architectures, pretraining data, and IR adaptation strategies: OpenAI-CLIP-B32, OpenAI-CLIP-L14~\cite{radford2021}, OpenCLIP-B32~\cite{openclip}, RemoteCLIP-B32~\cite{remoteclip}, and GeoRSCLIP-B32~\cite{georsclip}. For VLM evaluation, we use six state-of-the-art models: Qwen2.5-VL-7B~\cite{qwen25vl}, InstructBLIP~\cite{instructblip}, LLaVA-1.5 (7B)~\cite{liu2023llava}, LLaVA-1.6 (7B)~\cite{liu2024llava}, GeoChat (7B)~\cite{geochat}, and H2RSVLM~\cite{h2rsvlm}.

\textbf{Datasets.} Our dataset is constructed from five public RS sources (NWPU-Caption~\cite{nwpucaption}, RSICD~\cite{rsicd}, RSITMD~\cite{rsitmd}, RS5M~\cite{rs5m}, SkyScript~\cite{skyscript}), filtered to infrared samples to prevent RGB information leakage. The filtering procedure removes samples whose image filenames or metadata contain RGB indicators. Final class distribution and per-source proportions are detailed in the supplementary material. The test split contains 10{,}000 IR image-text pairs, of which 9{,}720 carry a remote-sensing scene-category label and are used for the CLIP zero-shot classification attack; the validation split (used for ablations) contains 416 samples. The VLM evaluation uses 1{,}000 randomly sampled labeled IR images with annotations for four tasks.

\textbf{Metrics.} For CLIP, each backbone is a zero-shot scene classifier over remote-sensing scene categories: each IR image is assigned the scene class whose text prompt has the highest image--text cosine similarity. We report Attack Success Rate (ASR): the percentage of samples whose adversarial top-1 scene prediction differs from the clean top-1 prediction.\footnote{ASR is a paired flip rate---clean vs.\ adversarial prediction on the same image---so the clean condition is the reference (0\% by construction) and no separate clean-accuracy baseline is required. The denominator is the full evaluated set.} For the cross-model transfer study (\cref{sec:transfer_results}), where a single perturbation is applied to held-out samples and evaluated by nearest-caption retrieval over a 1,000-sample candidate pool, we additionally report retrieval top-1 flip rate, confidence drop, and the number of unique adversarial top-1 captions as an error-diversity measure. For VLM evaluation, we report ROUGE-L for caption quality, scene accuracy, object F1, and IR-cue accuracy. All metrics use the full evaluation set. For VLM experiments, we assess every clean-vs-attack change with a two-proportion test ($n{=}1000$) and Wilson 95\% confidence intervals, correcting 18 comparisons with both Bonferroni and Benjamini--Hochberg (FDR) procedures. Because clean and adversarial predictions are evaluated on the same images, this unpaired test treats them as independent and is therefore conservative relative to an exact paired (McNemar) test; the significance markers we report thus understate rather than overstate the effects.

\textbf{Baselines.} We compare against four IR-specific physical attack methods: (1)~Atmospheric thermal turbulence---adding synthetic turbulence patterns generated by phase-screen propagation models~\cite{mao2021turbulence} (distinct from our airflow template bank, which uses randomized heat-kernel convolution, so that the baseline is an independent turbulence formulation rather than a weakened version of our prior); (2)~IR fixed-pattern stripe noise---adding periodic column-wise noise mimicking sensor readout artifacts; (3)~Thermal hot/cold blocks---overlaying rectangular regions with elevated or depressed temperatures, following the HOTCOLD block design~\cite{hotcold}; (4)~Thermal sensor drift/non-uniformity---applying pixel-wise gain and offset variations simulating detector degradation. All baselines operate under the same $\varepsilon = 100$ constraint. We deliberately restrict the comparison to IR-specific physical attacks, which share AirflowAttack's physical-plausibility motivation; unstructured digital baselines (pixel-space UAP, Gaussian noise) address the orthogonal question of whether perturbation \emph{structure} matters, which we examine directly through the loss-component and spatial-position ablations in \cref{sec:ablations}.

\subsection{Experiment 1: Zero-Shot Scene Classification Attack}
\label{sec:clip_results}

\begin{figure}[t]
  \centering
  \includegraphics[width=\textwidth]{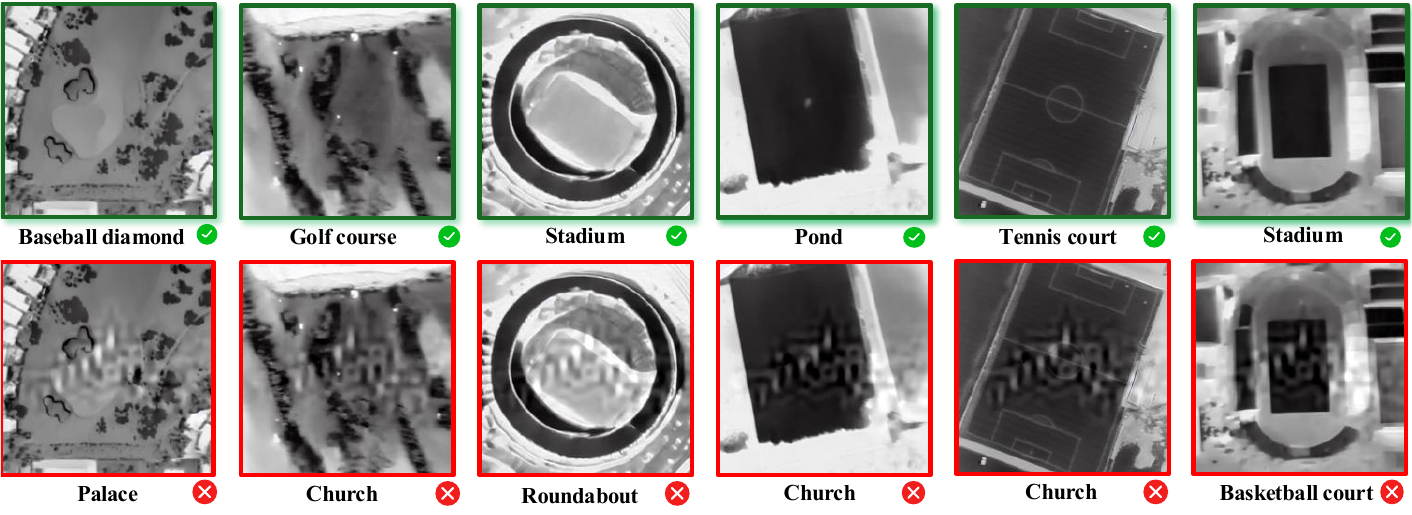}
  \caption{Qualitative zero-shot scene classification under AirflowAttack on six IR images. \textbf{Top row (green):} clean inputs correctly classified by the surrogate CLIP model. \textbf{Bottom row (red):} the same images with the universal thermal-airflow perturbation, now misclassified as unrelated categories. The faint, coherent airflow-like texture flips top-1 predictions while preserving human-recognizable scene content.}
  \label{fig:zeroshot}
\end{figure}

\Cref{fig:zeroshot} shows representative zero-shot classification flips, and \cref{tab:clip_comparison} reports the ASR of AirflowAttack compared to four IR-specific baselines across five CLIP backbones. AirflowAttack achieves the highest ASR on every backbone, with a mean ASR of 48.5\% versus 33.6\% (turbulence), 37.0\% (stripe noise), 29.2\% (hot/cold blocks), and 27.7\% (sensor drift). The strongest result is on OpenAI-CLIP-B32 (54.2\%), the surrogate model, but high ASR persists on architecturally distinct models: GeoRSCLIP-B32 (51.7\%), RemoteCLIP-B32 (50.6\%), and OpenCLIP-B32 (47.9\%). Even on OpenAI-CLIP-L14, which uses a different visual encoder (ViT-L/14 vs.\ ViT-B/32), AirflowAttack achieves 38.3\% ASR, exceeding the best baseline (turbulence at 36.9\%). Because ASR is measured as a paired flip of the top-1 scene prediction relative to clean, it isolates the effect of the perturbation itself, independent of each backbone's underlying clean accuracy.

\begin{table}[t]
  \caption{CLIP zero-shot scene-classification attack success rate (ASR \%) across five backbones at $\varepsilon=100$. ASR is the fraction of test samples whose top-1 predicted scene class under attack differs from the clean prediction. AirflowAttack exceeds all four IR-specific physical baselines on every backbone.}
  \label{tab:clip_comparison}
  \centering
  \resizebox{\textwidth}{!}{%
  \begin{tabular}{@{}lccccc@{}}
    \toprule
    Method & OpenAI-CLIP-L14 & OpenAI-CLIP-B32 & OpenCLIP-B32 & RemoteCLIP-B32 & GeoRSCLIP-B32 \\
    \midrule
    Atmospheric turbulence & 36.9 & 35.6 & 36.1 & 28.8 & 30.8 \\
    Stripe noise & 28.2 & 39.5 & 38.3 & 41.3 & 37.8 \\
    Hot/cold blocks & 25.2 & 29.6 & 28.5 & 31.8 & 30.7 \\
    Sensor drift & 24.9 & 32.1 & 25.9 & 27.9 & 27.7 \\
    \midrule
    \rowcolor{oursrow} AirflowAttack (Ours) & \textbf{38.3} & \textbf{54.2} & \textbf{47.9} & \textbf{50.6} & \textbf{51.7} \\
    \bottomrule
  \end{tabular}%
  }
\end{table}

Two observations merit emphasis. First, the ranking of baselines is inconsistent across backbones: stripe noise is the strongest baseline on four backbones (OpenAI-CLIP-B32, OpenCLIP-B32, RemoteCLIP-B32, and GeoRSCLIP-B32), while turbulence leads only on OpenAI-CLIP-L14. AirflowAttack is the only method that consistently dominates. Second, the performance gap between AirflowAttack and baselines is largest precisely on the backbones most relevant to IR deployment---RemoteCLIP-B32 and GeoRSCLIP-B32---where domain-specific pretraining appears to amplify sensitivity to physically structured perturbations.

\subsubsection{Cross-Model Transfer.}
\label{sec:transfer_results}

A defining property of a universal perturbation is cross-model transfer: a perturbation optimized on one model should remain effective on others without target-model gradients or adaptation. Using a single UAP optimized on OpenAI-CLIP-B32 and a nearest-caption retrieval probe over a 1,000-sample candidate pool, the perturbation changes the top-1 nearest caption for 94.4\% (OpenAI-CLIP-L14) to 98.8\% (GeoRSCLIP-B32) of held-out samples across all five backbones, despite being optimized solely on the surrogate. This indicates that the learned airflow pattern does not merely exploit idiosyncratic surrogate decision boundaries, but perturbs thermal representations shared by general and remote-sensing CLIP variants. This retrieval flip rate is a more sensitive probe than the scene-classification ASR of \cref{tab:clip_comparison}---any change in the nearest caption counts, not only a change of predicted scene class---so its values are correspondingly higher and are not directly comparable. Notably, a higher confidence drop does not imply a higher flip rate: OpenAI-CLIP-L14 shows the largest confidence reduction yet the lowest flip rate, which may indicate wider retrieval margins in the ViT-L/14 embedding space, though we do not test this directly. Per-model flip rates, confidence drops, and caption-diversity statistics are reported in \cref{tab:transfer}.

\begin{table}[t]
  \caption{Cross-model transfer of one AirflowAttack UAP optimized on OpenAI-CLIP-B32 and applied without target access to 1,000 held-out samples per backbone. Flip rate (\%) measures top-1 nearest-caption changes relative to clean; Conf.\ Drop is the mean cosine-similarity reduction to the correct caption. The UAP transfers strongly across architectures (94.4--98.8\%); ViT-L/14 is most resistant despite the largest confidence drop, suggesting wider retrieval margins.}
  \label{tab:transfer}
  \centering
  \resizebox{\textwidth}{!}{%
  \begin{tabular}{@{}lccccc@{}}
    \toprule
    Metric & OpenAI-CLIP-L14 & OpenAI-CLIP-B32 & OpenCLIP-B32 & RemoteCLIP-B32 & GeoRSCLIP-B32 \\
    \midrule
    Flip Rate (\%) & 94.4 & 98.1 & 97.3 & 98.5 & 98.8 \\
    Conf.\ Drop & 0.223 & 0.149 & 0.176 & 0.077 & 0.165 \\
    \bottomrule
  \end{tabular}%
  }
\end{table}

\begin{table*}[h]
\centering
\caption{Captioning impact across six downstream VLMs measured by ROUGE-L. All values are percentages. Lower values indicate stronger caption degradation. The best attack result in each column is highlighted.}
\label{tab:caption_rougel}
\setlength{\tabcolsep}{5pt}
\renewcommand{\arraystretch}{1.08}
\resizebox{\textwidth}{!}{%
\begin{tabular}{lcccccc}
\toprule
Method & Qwen2.5-VL & InstructBLIP & LLaVA-1.5 & LLaVA-1.6 & GeoChat & H2RSVLM \\
\midrule
Hot/Cold Block & 11.75 & 12.31 & 10.59 & 10.19 & 10.13 & 11.69 \\
IR Stripe & 11.80 & 12.29 & 10.58 & \cellcolor{bestgreen}\textbf{10.11} & 10.07 & 11.65 \\
Thermal Drift & 11.77 & 12.13 & 10.41 & 10.20 & 10.08 & \cellcolor{bestgreen}\textbf{11.59} \\
Thermal Turb. & 11.74 & 12.11 & \cellcolor{bestgreen}\textbf{10.37} & 10.16 & 9.84 & 11.74 \\
\cmidrule(lr){1-7}
\textbf{Ours} & \cellcolor{bestgreen}\textbf{11.70} & \cellcolor{bestgreen}\textbf{11.91} & \textbf{10.42} & \textbf{10.24} & \cellcolor{bestgreen}\textbf{9.80} & \textbf{11.64} \\
\bottomrule
\end{tabular}%
}
\end{table*}

\begin{figure*}[t]
  \centering
  \includegraphics[width=\textwidth]{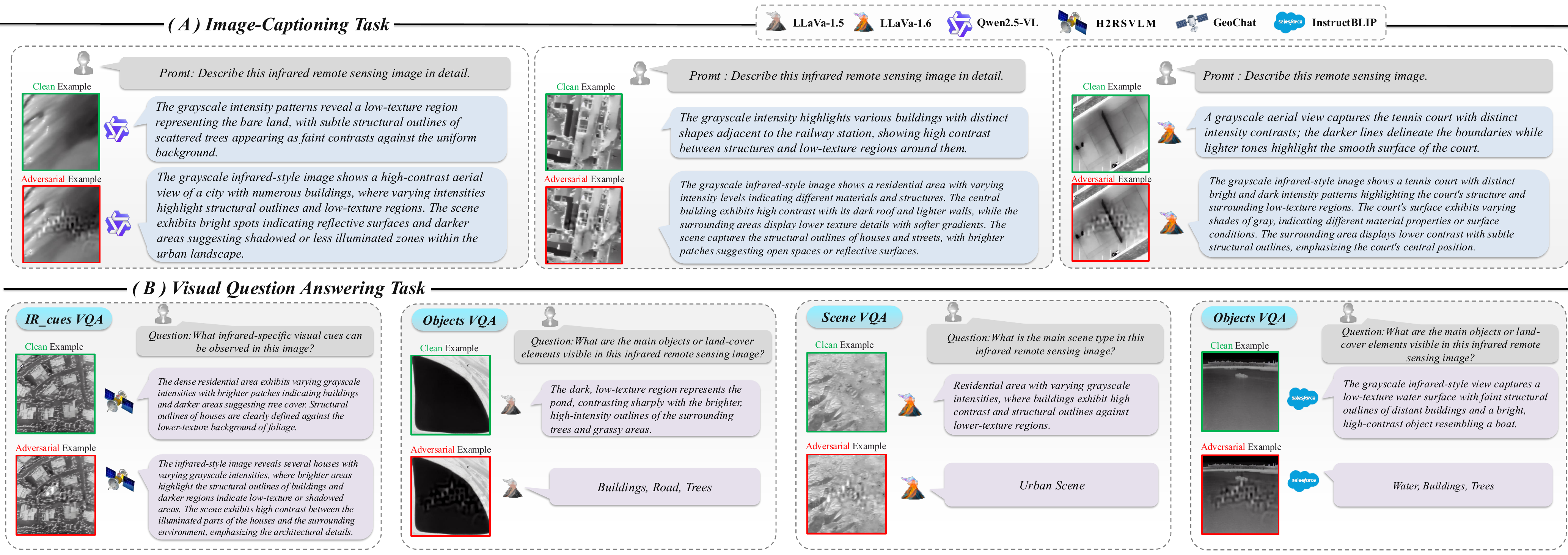}
  \caption{Effect of AirflowAttack on VLM captioning and visual question answering. Under attack, models produce more generic captions and confabulate thermal cues (temperature gradients, convection signatures) absent from the clean image, illustrating the IR-cue paradox quantified in \cref{tab:vqa_transfer}.}
  \label{fig:vqacaption}
\end{figure*}

\subsection{Experiment 2: Image Captioning Attack}
\label{sec:vlm_caption}

We next transfer the AirflowAttack UAP---the same perturbation optimized on the CLIP surrogate---to six generative VLMs, applied identically to all inputs. We first evaluate infrared image captioning, scoring the generated caption against the reference with ROUGE-L. \Cref{tab:caption_rougel} reports ROUGE-L under AirflowAttack and the four physical baselines for all six models.

Caption quality shows a nuanced pattern. AirflowAttack attains the lowest ROUGE-L on three of the six models (Qwen2.5-VL, InstructBLIP, and GeoChat), matching or exceeding the physical baselines elsewhere. The absolute changes, however, are small and non-monotonic across methods: ROUGE-L is a lexical-overlap metric rather than a measure of factual accuracy, so a perturbation that induces more generic, repetitive captions can raise n-gram overlap even as semantic content degrades. Qualitative inspection (\cref{fig:vqacaption}) confirms that attacked captions frequently describe non-existent objects or misidentify scene categories; a lexical-overlap metric therefore understates the true degradation, motivating the reasoning-based VQA evaluation in Experiment~3.

\subsection{Experiment 3: Visual Question Answering Attack}
\label{sec:vlm_vqa}

We then probe higher-level reasoning through three VQA-style tasks: scene classification (accuracy, over a VLM-specific label set distinct from the CLIP zero-shot categories in Experiment~1), object presence recognition (F1), and infrared-cue explanation (accuracy). \Cref{tab:vqa_transfer} reports clean, physical-baseline, and AirflowAttack scores.

\begin{table*}[t]
\centering
\caption{VQA impact across six downstream VLMs. Scene Acc., Object F1, and IR-Cue Acc. are reported in percentages. Lower adversarial values indicate stronger degradation. The best attack result in each column is highlighted.}
\label{tab:vqa_transfer}
\setlength{\tabcolsep}{4.2pt}
\renewcommand{\arraystretch}{1.08}
\resizebox{\textwidth}{!}{%
\begin{tabular}{>{\centering\arraybackslash}m{1.6cm}lcccccc}
\toprule
Metric & Method & Qwen2.5-VL & InstructBLIP & LLaVA-1.5 & LLaVA-1.6 & GeoChat & H2RSVLM \\
\midrule
\multirow{6}{1.6cm}{\centering Scene Acc.}
& Clean & 35.24 & 23.66 & 21.46 & 23.54 & 24.27 & 24.27 \\
& Hot/Cold Block & 21.83 & 19.51 & 17.68 & 19.15 & 21.10 & 21.83 \\
& Thermal Drift & 21.95 & 17.93 & 17.07 & 19.15 & 20.49 & 22.20 \\
& IR Stripe & 20.85 & 17.56 & 16.71 & 18.78 & 20.73 & 21.59 \\
& Thermal Turb. & 21.59 & 16.59 & 14.39 & 16.71 & 19.76 & 18.78 \\
\cmidrule(lr){2-8}
& \textbf{Ours} & \textbf{21.83} & \textbf{14.63} & \textbf{16.22} & \textbf{16.22} & \textbf{19.39} & \textbf{20.37} \\
\midrule
\multirow{6}{1.6cm}{\centering Object F1}
& Clean & 33.19 & 24.18 & 26.72 & 25.35 & 25.91 & 25.52 \\
& Thermal Drift & 28.12 & 24.59 & 24.57 & 24.87 & 22.65 & 23.86 \\
& Hot/Cold Block & 27.69 & 23.83 & 24.77 & 25.07 & 23.04 & 23.91 \\
& IR Stripe & 27.21 & 23.14 & 24.29 & 25.18 & 23.08 & 22.79 \\
& Thermal Turb. & 26.90 & 22.56 & 24.23 & 23.61 & 22.34 & 22.24 \\
\cmidrule(lr){2-8}
& \textbf{Ours} & \textbf{28.60} & \textbf{23.64} & \textbf{24.47} & \textbf{25.28} & \textbf{21.73} & \textbf{22.53} \\
\midrule
\multirow{6}{1.6cm}{\centering IR-Cue Acc.}
& Clean & 86.59 & 81.13 & 98.80 & 95.34 & 98.13 & 92.40 \\
& IR Stripe & 93.90 & 84.00 & 96.31 & 90.84 & 95.71 & 89.09 \\
& Hot/Cold Block & 94.01 & 83.83 & 96.47 & 90.17 & 95.11 & 88.71 \\
& Thermal Drift & 94.06 & 83.29 & 95.96 & 89.93 & 95.51 & 88.71 \\
& Thermal Turb. & 94.07 & 83.11 & 94.66 & 89.86 & 95.74 & 88.57 \\
\cmidrule(lr){2-8}
& \textbf{Ours} & \textbf{94.43} & \textbf{82.94} & \textbf{97.44} & \textbf{91.90} & \textbf{95.70} & \textbf{88.84} \\
\bottomrule
\end{tabular}%
}
\end{table*}

\textbf{Scene classification} suffers the most severe and statistically robust degradation: the drops on Qwen2.5-VL-7B (35.24$\rightarrow$21.83, 38.1\%), InstructBLIP (23.66$\rightarrow$14.63, 38.2\%), LLaVA-1.5 (21.46$\rightarrow$16.22) and LLaVA-1.6 (23.54$\rightarrow$16.22, 31.1\%) all survive Bonferroni correction ($p<0.0028$), and GeoChat survives FDR. The perturbation systematically misleads models about scene identity, a failure mode with direct operational consequences for IR-based surveillance and monitoring.

\textbf{Object recognition} (F1) shows a much weaker effect. Only Qwen2.5-VL-7B and GeoChat reach FDR significance, and \emph{no} object-F1 change survives Bonferroni correction; the changes on InstructBLIP ($p{=}0.75$), LLaVA-1.5 ($p{=}0.26$), LLaVA-1.6 (unchanged, $p{=}1.0$) and H2RSVLM ($p{=}0.12$) are statistically indistinguishable from zero at $n{=}1000$. We therefore do not claim broad object-recognition degradation: object presence is markedly more robust to the perturbation than scene identity, plausibly because it depends on localized high-contrast structure that a low-magnitude airflow field leaves intact.

\textbf{IR cue accuracy} reveals the most intriguing finding: two of six models (Qwen2.5-VL-7B and InstructBLIP) exhibit \emph{higher} IR-cue detection rates under attack, most strikingly Qwen2.5-VL-7B, whose accuracy rises from 86.59 to 94.43---the opposite of the intended attack effect. This occurs because the airflow perturbation introduces structured thermal patterns that these models interpret as genuine IR phenomena---temperature gradients, thermal shadows, convection signatures---triggering confident but incorrect IR-cue identifications. The perturbation effectively manufactures thermal ``evidence'' that the models' IR understanding modules latch onto, demonstrating a form of adversarial exploitation unique to the thermal modality. The remaining four models show only modest IR-cue decreases, so this confabulation effect is model-dependent rather than universal.

\begin{figure}[t]
  \centering
  \includegraphics[width=\textwidth]{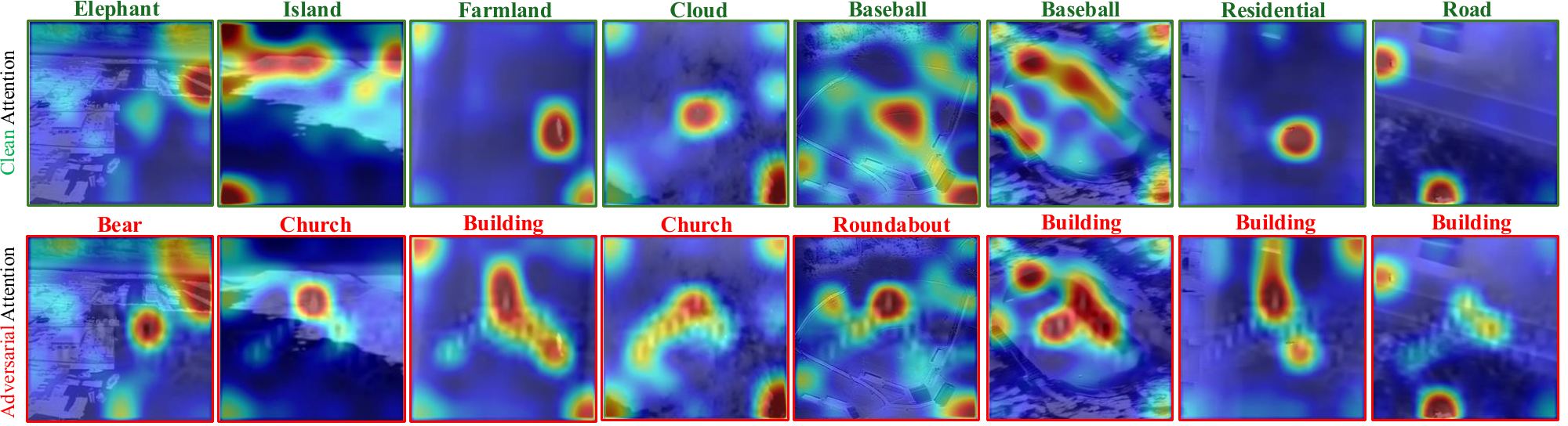}
  \caption{Attention shift under AirflowAttack. Top row: clean model attention (Grad-CAM) with the correct top-1 scene prediction; bottom row: attention on the same images under the perturbation, with the flipped adversarial prediction. The airflow perturbation redirects the model's spatial attention away from scene-defining content, driving the top-1 class change that ASR (\cref{tab:clip_comparison}) measures.}
  \label{fig:heatmap}
\end{figure}

To understand \emph{how} the perturbation induces these failures, \cref{fig:heatmap} visualizes the attention shift it causes: the perturbation redirects model attention away from scene-defining regions, flipping the top-1 prediction (e.g., island$\rightarrow$church, farmland$\rightarrow$building). We now analyze the factors that govern this effect through a series of ablations.

\subsection{Ablation Studies}
\label{sec:ablations}

We conduct extensive ablations to understand the factors governing AirflowAttack's efficacy. All ablations are performed on OpenAI-CLIP-B32 using the validation set, with the full attack configuration ($\varepsilon = 100$, latent dimension 32, residual ratio 0.60, 800 steps, $\eta = 0.055$, loss weights $\alpha = 8$, $\beta = 2$) as the reference.

\subsubsection{Perturbation Strength}
\label{sec:abl_eps}

\begin{figure}[t]
  \centering
  \begin{subfigure}[t]{0.49\textwidth}
    \centering
    \includegraphics[width=\textwidth]{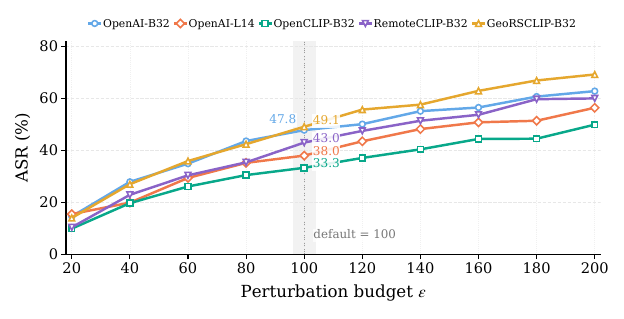}
    \caption{Perturbation strength $\varepsilon$.}
    \label{fig:abl_eps}
  \end{subfigure}
  \hfill
  \begin{subfigure}[t]{0.49\textwidth}
    \centering
    \includegraphics[width=\textwidth]{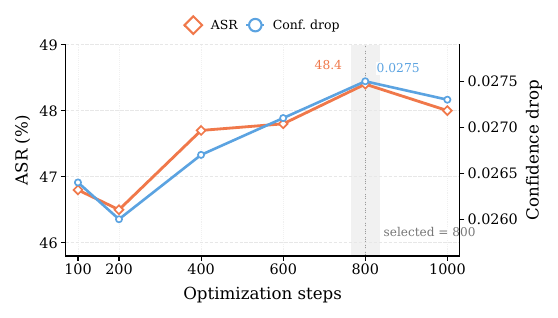}
    \caption{Optimization steps.}
    \label{fig:abl_steps}
  \end{subfigure}
  \caption{Optimization ablations on OpenAI-CLIP-B32 (validation set). (a) ASR rises monotonically with the perturbation budget $\varepsilon$ across all five backbones; the dashed line marks our reference $\varepsilon=100$. (b) ASR converges by $\sim$800 steps and is stable thereafter, indicating robustness to early stopping.}
  \label{fig:abl_optim}
\end{figure}

As shown in \cref{fig:abl_eps}, ASR increases monotonically with $\varepsilon$ across all five CLIP backbones: 12.9\% at $\varepsilon = 20$, 42.2\% at $\varepsilon = 100$ (our reference budget), and 59.7\% at $\varepsilon = 200$ (mean across backbones). Note that ablation ASR values (42.2\% at $\varepsilon=100$) differ from \cref{tab:clip_comparison}'s ASR (48.5\% mean at $\varepsilon=100$) because ablations are run on a smaller validation subset (416 samples) rather than the full 9,720-sample test set. The near-linear relationship between perturbation budget and attack success suggests that the airflow perturbation remains effective across a wide range of visibility levels. At $\varepsilon = 20$, the perturbation is nearly invisible (less than 8\% of the dynamic range), yet still achieves non-trivial ASR, indicating that even subtle thermal patterns can disrupt IR scene understanding.

\subsubsection{Loss Component Analysis}
\label{sec:abl_loss}

\Cref{fig:abl_loss_fig} decomposes the contribution of each loss term. The dominant factor is clearly the confidence loss: the configurations that include it (full at 47.9\% and no-air at 48.0\%) reach $\sim$48\% ASR, whereas the two configurations driven primarily by the airflow prior (no-conf and fixed-prior) reach only 39.5\% and 38.5\%. This $\sim$9-point gap confirms that $\mathcal{L}_{\text{conf}}$ is the primary driver of attack efficacy, while $\mathcal{L}_{\text{air}}$ governs physical plausibility (airflow correlation rises toward 0.985 as its weight increases).

Within the confidence-dominated regime, the effect of adding the airflow term is negligible: the full objective (47.9\%) and the no-air variant (48.0\%) differ by only 0.1 points---well within the run-to-run noise floor at $n{=}416$ validation samples. Adding the airflow prior thus does \emph{not} measurably reduce ASR, while it substantially raises physical plausibility (correlation $0.844\rightarrow0.893$). We therefore include $\mathcal{L}_{\text{air}}$ as an essentially cost-free way to make the perturbation physically interpretable, rather than claiming it improves attack strength.

\subsubsection{Spatial Position}
\label{sec:abl_position}

\begin{figure}[t]
  \centering
  \begin{subfigure}[t]{0.49\textwidth}
    \centering
    \includegraphics[width=\textwidth]{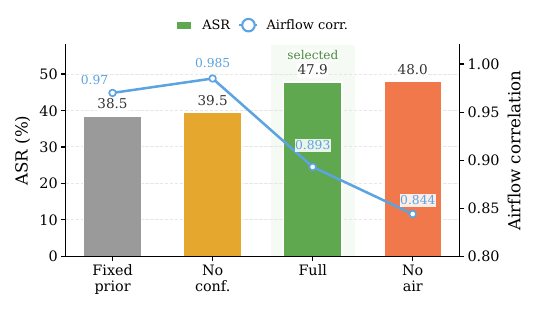}
    \caption{Loss components.}
    \label{fig:abl_loss_fig}
  \end{subfigure}
  \hfill
  \begin{subfigure}[t]{0.49\textwidth}
    \centering
    \includegraphics[width=\textwidth]{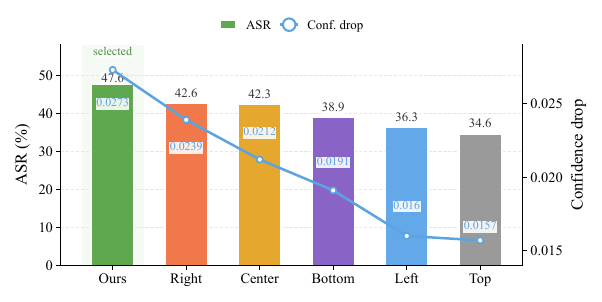}
    \caption{Spatial position.}
    \label{fig:abl_pos_fig}
  \end{subfigure}
  \caption{Loss-composition and spatial-position ablations on OpenAI-CLIP-B32 (validation set). (a) ASR (bars) is driven by the confidence loss, while airflow correlation (line) rises as the airflow prior is weighted more heavily. (b) ASR when the perturbation is restricted to one image region: covering the full image is most effective, followed by the right and center regions that hold scene-defining content.}
  \label{fig:abl_losspos}
\end{figure}

The perturbation's spatial location strongly affects performance. We partition it into six regions (full image, top, bottom, left, right, center) and restrict $\delta$ to each region with zero-padding elsewhere, as shown in \cref{fig:abl_pos_fig}. The full image reaches 47.6\% ASR, followed by right (42.6\%), center (42.3\%), bottom (38.9\%), left (36.3\%), and top (34.6\%). Right and center regions are most impactful, possibly because IR remote-sensing scenes often place key content such as buildings, vehicles, and terrain features near central or right-side areas, although we do not verify this attribution. The 13-point gap between full-image and top-only perturbations indicates that AirflowAttack depends on covering regions used for scene understanding.

\subsubsection{Hyperparameter Sensitivity}
\label{sec:abl_hyper}

We evaluate sensitivity to key hyperparameters, each swept independently with the others held at the reference configuration. \textbf{Learning rate}: ASR is stable in the 47.5--48.5\% range for $\eta \in [0.04, 0.07]$, optimal at $\eta = 0.055$. \textbf{Optimization steps}: ASR ranges 47.7--48.4\% for 400--1000 steps and peaks at 800 (\cref{fig:abl_steps}), indicating rapid convergence and robustness to early stopping. \textbf{Latent dimension} and \textbf{residual ratio}: within each sweep, a latent code of dimension 32 and a residual ratio $r = 0.60$ are the best-performing settings, and are adopted as the reference. Across all four hyperparameters, ASR varies by less than two points within the ranges tested, demonstrating that AirflowAttack is robust to hyperparameter variation.

\section{Conclusion}
\label{sec:conclusion}

We presented AirflowAttack, to the best of our knowledge the first adversarial attack for infrared remote-sensing vision-language models and the first to repurpose thermal-airflow turbulence as an adversarial perturbation. A single physically interpretable, input-agnostic perturbation achieves a mean zero-shot classification ASR of 48.5\% (vs.\ 27.7--37.0\% for IR-specific baselines), transfers from one surrogate to five CLIP backbones at a 94.4--98.8\% flip rate, and---applied to six VLMs---significantly degrades scene classification while, on some models, paradoxically raising IR-cue confidence through adversarial confabulation. The ablations further show that attack strength is primarily driven by the confidence loss, whereas the airflow prior improves physical plausibility with negligible ASR cost. Together with a benchmark spanning eleven models and four tasks, these findings establish that IR remote-sensing VLMs are not robust by default and that physically grounded thermal perturbations constitute a potent, modality-specific threat vector, motivating future work on detection, purification, and adversarial training for IR-specific defenses.

% ---- Bibliography ----
\bibliographystyle{splncs04}
\bibliography{main}

\clearpage
\appendix
\setcounter{figure}{0}
\setcounter{table}{0}
\renewcommand{\thefigure}{\Alph{section}.\arabic{figure}}
\renewcommand{\thetable}{\Alph{section}.\arabic{table}}
\renewcommand{\theHsection}{appendix.\Alph{section}}
\renewcommand{\theHsubsection}{appendix.\Alph{section}.\arabic{subsection}}
\renewcommand{\theHfigure}{appendix.\Alph{section}.\arabic{figure}}
\renewcommand{\theHtable}{appendix.\Alph{section}.\arabic{table}}
\setcounter{algocf}{0}
\renewcommand{\thealgocf}{A.\arabic{algocf}}
\renewcommand{\theHalgocf}{A.\arabic{algocf}}

\section{Detailed Method Formulation}
\label{app:method}

This appendix expands the components abbreviated in the method section of the
main paper. We give the full perturbation parameterization
(Appendix~\ref{app:generator}), the airflow prior field and its spatial gate
(Appendices~\ref{app:templates}--\ref{app:gate}), the exact form of the airflow
prior and confidence losses (Appendices~\ref{app:aircorr}--\ref{app:conf}), and
the combined objective and optimization details (Appendix~\ref{app:optim}).
Throughout, $H\times W$ is the IR input resolution,
$x\in\mathbb{R}^{H\times W}$ a single-channel thermal image, and
$\delta\in\mathbb{R}^{H\times W}$ the universal perturbation. Images are
represented in the $[0,1]$ intensity range during optimization; when we write
$\varepsilon=100$ in tables, the corresponding value used in equations and code
is $100/255$.

\subsection{Generator Parameterization}
\label{app:generator}

Rather than optimizing $\delta$ freely in pixel space, we parameterize it as a
composition of a fixed airflow \emph{prior} field and a low-dimensional
learnable \emph{residual}, so that the optimizer explores only a compact,
physically anchored subspace. Let $P\in\mathbb{R}^{H\times W}$ denote the unit-
normalized airflow prior (its construction is given in
Appendix~\ref{app:templates}) and $G\in[0,1]^{H\times W}$ a spatial gate derived
from it (Appendix~\ref{app:gate}). The two learnable quantities are (i) a coarse
latent map $z\in\mathbb{R}^{d\times d}$ ($d=32$) and (ii) a scalar amplitude
$a=0.85+0.30\,\operatorname{sigmoid}(\rho)$ with learnable logit $\rho$. The latent is
decoded into a full-resolution residual by bicubic upsampling followed by a
Gaussian blur $\mathcal{B}$ (kernel $9$, $\sigma=2.5$), then mean-removed and
max-normalized:
\begin{equation}
R \;=\; \operatorname{norm}\!\big(\mathcal{B}(\mathcal{U}(z))\big)\odot G,
\qquad
\operatorname{norm}(u)=
\frac{u-\operatorname{mean}(u)}
{\max_i|u_i-\operatorname{mean}(u)|+\xi},\quad \xi=10^{-6}.
\label{eq:residual}
\end{equation}
The perturbation is then formed by an amplitude-scaled, tanh-squashed
combination of prior and residual, gated and projected to the $L_\infty$ ball:
\begin{equation}
\delta \;=\; \operatorname{clip}\!\Big(\varepsilon\,\tanh\!\big(a\,P + r\,R\big)\odot G,\;-\varepsilon,\;\varepsilon\Big),
\label{eq:gen_out}
\end{equation}
where $\odot$ is elementwise product and $r$ is the \emph{residual scale}.
Setting $r=0.60$ lets the learnable residual reshape the prior substantially
while the airflow term $a\,P$ keeps the pattern physically anchored; this
empirically balances attack strength against physical plausibility (see the main
paper's ablation study). Optimization thus updates only the latent $z$ and the
amplitude logit $\rho$---$d^2+1=325$ parameters in the reference
configuration---rather than all $HW$ pixels, which regularizes the perturbation
toward smooth, spatially coherent airflow structure and shrinks the search
space.

\subsection{Airflow Prior Field}
\label{app:templates}

The airflow prior $P$ is a single, fixed thermal-airflow field that encodes the
spatially correlated temperature fluctuations produced when turbulent airflow
mixes air parcels of differing temperature. It is precomputed once as a
signed single-channel field, resized to the model input resolution $H\times W$,
and unit-normalized by its peak magnitude:
\begin{equation}
P \;=\; \frac{\tilde P}{\max_i|\tilde P_i|},
\label{eq:prior_norm}
\end{equation}
where $\tilde P$ is the raw airflow field loaded from the precomputed asset.
Because the prior is fixed throughout optimization and the perturbation is
anchored to it through \eqref{eq:gen_out}, the attack explores airflow-like
patterns rather than arbitrary noise. This fixed-prior design is deliberately
distinct from the phase-screen propagation model used for the
\emph{atmospheric-turbulence baseline} in the main paper, so that the baseline
is an independent turbulence formulation rather than a weakened variant of our
own prior.

\subsection{Spatial Gate}
\label{app:gate}

To keep the perturbation concentrated on the structured regions of the prior
and suppress it elsewhere, we derive a soft spatial gate $G\in[0,1]^{H\times W}$
from the normalized prior by soft-thresholding its magnitude at $\gamma$
(gate threshold), blurring with a Gaussian kernel of size 21 and $\sigma=5.0$,
and optionally masking to a spatial region:
\begin{equation}
G \;=\; \operatorname{norm}_{\max}\!\Big(\mathcal{B}\big(\operatorname{clip}(\tfrac{|P|-\gamma}{1-\gamma},0,1)\big)\odot \Pi\Big),
\label{eq:gate}
\end{equation}
where $\mathcal{B}$ is the Gaussian blur, $\Pi$ is a position mask
($\Pi\equiv 1$ for the default full-image setting; the spatial-position ablation
varies it over the top, bottom, left, right, and center regions), and
$\operatorname{norm}_{\max}$ rescales to a peak of one. The gate multiplies both the residual
(\eqref{eq:residual}) and the final perturbation (\eqref{eq:gen_out}), so energy
is spent only where the airflow prior is salient.

\subsection{Airflow Prior Loss}
\label{app:aircorr}

The airflow prior loss keeps the optimized perturbation aligned with the prior
in shape. Writing $\hat u = u-\operatorname{mean}(u)$ for a mean-removed field
and $\delta_{\mathrm{g}}=\delta\odot G$ for the gated perturbation, we use a
cosine alignment between $\delta_{\mathrm{g}}$ and the prior $P$:
\begin{equation}
\mathcal{L}_{\text{air}}
\;=\; 1-\frac{\langle \hat\delta_{\mathrm{g}},\,\hat P\rangle}
                {(\lVert\hat\delta_{\mathrm{g}}\rVert_2+\xi)(\lVert\hat P\rVert_2+\xi)},
\label{eq:air_full}
\end{equation}
which is one minus the spatial Pearson correlation between $\delta_{\mathrm{g}}$
and $P$ and is invariant to the perturbation's overall scale and offset. The
\emph{airflow correlation} reported in the ablations is exactly
$1-\mathcal{L}_{\text{air}}$ at the converged perturbation. We also implemented
a candidate frequency-domain prior $\mathcal{L}_{\text{spec}}$, defined as a
Jensen--Shannon divergence between the $\ell_1$-normalized $|\mathrm{FFT}|^2$
maps of $\delta_{\mathrm{g}}$ and $P$. This spectral term is reported only as an
ablation/candidate component; it is not active in the final reference setting.

\subsection{Confidence Loss}
\label{app:conf}

The active attack objective operates in the surrogate CLIP classifier induced by
the scene-prompt set. Let $f_I(\cdot)$ and $f_T(\cdot)$ be the $\ell_2$-
normalized image and text encoders, let
$\operatorname{sim}(u,v)=f_I(u)^\top f_T(v)$, and let $\alpha$ be the CLIP logit
scale used by the checkpoint. For a scene class $c$ with prompt text
$\pi(c)$, the zero-shot logit is
\begin{equation}
\ell_c(x,\delta)=
\alpha\,\operatorname{sim}\!\big(f_I(\operatorname{pre}(x+\delta)),f_T(\pi(c))\big),
\label{eq:clip_logit}
\end{equation}
where $\operatorname{pre}(\cdot)$ denotes the model's native resizing and
normalization after the image has been clipped to the valid intensity range.
The clean class is the model's own top-1 prediction
$y=\arg\max_c \ell_c(x,0)$, not an external annotation. We define
\begin{equation}
c_y=\operatorname{softmax}(\ell(x,0))_y,\qquad
p_y(\delta)=\operatorname{softmax}(\ell(x,\delta))_y .
\label{eq:clean_adv_prob}
\end{equation}
The reference attack minimizes the ratio $p_y(\delta)/c_y$, so the optimized
perturbation is explicitly driven to reduce the surrogate's confidence in the
clean top-1 decision. This choice matches the ASR definition used in evaluation:
a success is a paired clean-top1 to adversarial-top1 flip on the same image. It
is a prediction-stability attack and should not be read as a claim that the
clean top-1 class is always the human ground-truth label.

\subsection{Combined Objective and Optimization}
\label{app:optim}

The two headline losses of the main paper---an attack loss and the airflow
prior loss---are implemented through a small set of candidate terms, all
functions of the perturbation $\delta$ built by \eqref{eq:gen_out}. The
reference configuration activates the confidence-ratio term and the airflow
prior term; the margin, diversity, spectral, smoothness, and amplitude terms are
kept in the formulation because they are used in ablations or candidate
variants, but their reference weights are zero. For a paired caption $t^+$ and
batch negatives $\mathcal{T}^-$, let $s_t$ denote the image--text similarity to
caption $t$ under the adversarial image. The candidate terms are
\begin{align}
\mathcal{L}_{\text{conf}} &= \operatorname{mean}\big(p_y(\delta)/c_y\big),
\label{eq:conf_ratio}\\
\mathcal{L}_{\text{margin}} &= \tau_m\,\operatorname{mean}\;\operatorname{softplus}\!\Big(\tfrac{1}{\tau_m}\big(s_{t^+}-\max_{t\neq t^+}s_t+\kappa\big)\Big),
\label{eq:margin}
\end{align}
where minimizing $\mathcal{L}_{\text{conf}}$ directly reduces the probability of
the clean top-1 class defined in \eqref{eq:clean_adv_prob}. The optional margin
term drives the paired caption below the strongest in-batch competitor when its
weight is nonzero ($\kappa,\tau_m$ small constants). A candidate batch-diversity term
$\mathcal{L}_{\text{div}}$ (Jensen--Shannon divergence between the batch-mean
adversarial prediction and the clean prior) is used only in ablations to test
whether prediction collapse occurs. The active physical-plausibility term is the
airflow prior loss $\mathcal{L}_{\text{air}}$ \eqref{eq:air_full} and its
candidate spectral counterpart $\mathcal{L}_{\text{spec}}$
(Appendix~\ref{app:aircorr}). Two additional candidate regularizers control the
residual and the amplitude: a smoothness term
$\mathcal{L}_{\text{smooth}}=\mathrm{TV}(R)+\tfrac12\,\lVert\Delta R\rVert$
(total variation plus Laplacian of the residual) and a mean-budget term
$\mathcal{L}_{\text{amp}}=\operatorname{relu}\!\big(\bar{|\delta|}/\varepsilon-\mu\big)^2$
that penalizes mean perturbation magnitude above a budget $\mu$ when enabled. The full
objective is
\begin{equation}
\begin{aligned}
\mathcal{L}=\;&w_{\text{conf}}\mathcal{L}_{\text{conf}}+w_{\text{margin}}\mathcal{L}_{\text{margin}}+w_{\text{div}}\mathcal{L}_{\text{div}}+w_{\text{air}}\mathcal{L}_{\text{air}}\\
&+\,w_{\text{spec}}\mathcal{L}_{\text{spec}}+w_{\text{smooth}}\mathcal{L}_{\text{smooth}}+w_{\text{amp}}\mathcal{L}_{\text{amp}},
\end{aligned}
\label{eq:total_full}
\end{equation}
with the reference weights stated here. In the final setting,
$w_{\text{conf}}{=}8$ and $w_{\text{air}}{=}1$, while all other weights
($w_{\text{margin}},w_{\text{div}},w_{\text{spec}},w_{\text{smooth}},w_{\text{amp}}$)
are set to zero. The loss-ablation study of the main paper corresponds to changing these
weights one at a time (e.g.\ the \emph{no-air} variant sets
$w_{\text{air}}{=}0$; \emph{attack-only} keeps only the confidence-ratio term).

Supplementary Algorithm~\ref{alg:airflow_full} gives the exact implemented
update rule, refining the schematic procedure in the main paper.

\begin{algorithm}[H]
\small
\SetAlgoLined
\DontPrintSemicolon
\SetCommentSty{textnormal}
\KwIn{Surrogate CLIP encoders $f_I,f_T$; IR image--caption pairs; airflow prior asset $\tilde P$; budget $\varepsilon$, steps $T$, rate $\eta$; gate threshold $\gamma$, residual scale $r$, blur $\mathcal{B}$; loss weights $\{w_\bullet\}$}
\KwOut{Universal perturbation $\delta^\star$}
$P \leftarrow \tilde P/\max_i|\tilde P_i|$\tcp*{load and unit-normalize fixed prior}
$G \leftarrow \operatorname{norm}_{\max}\!\big(\mathcal{B}(\operatorname{clip}(\tfrac{|P|-\gamma}{1-\gamma},0,1))\odot\Pi\big)$\tcp*{spatial gate}
$z \leftarrow \mathbf{0}\in\mathbb{R}^{d\times d}$;\quad $\rho \leftarrow 0$\tcp*{residual latent, amplitude logit}
initialize AdamW over $(z,\rho)$ (default betas/epsilon, weight decay $10^{-4}$)\;
\For{$i = 1 \ldots T$}{
  set learning rate by 30-step warmup followed by cosine decay\;
  $a \leftarrow 0.85+0.30\,\operatorname{sigmoid}(\rho)$;\quad $R \leftarrow \operatorname{norm}(\mathcal{B}(\mathcal{U}(z)))\odot G$\tcp*{decode residual}
  $\delta \leftarrow \operatorname{clip}\big(\varepsilon\tanh(aP + rR)\odot G,\,-\varepsilon,\varepsilon\big)$\tcp*{form perturbation}
  sample image batch; $x^{\text{adv}} \leftarrow \operatorname{clip}(x+\delta,0,1)$ before model preprocessing\;
  $\mathcal{L} \leftarrow w_{\text{conf}}\mathcal{L}_{\text{conf}} + w_{\text{margin}}\mathcal{L}_{\text{margin}} + w_{\text{div}}\mathcal{L}_{\text{div}}$\;
  \hspace{1.2em}$+\, w_{\text{air}}\mathcal{L}_{\text{air}} + w_{\text{spec}}\mathcal{L}_{\text{spec}} + w_{\text{smooth}}\mathcal{L}_{\text{smooth}} + w_{\text{amp}}\mathcal{L}_{\text{amp}}$\tcp*{Eq.\ \eqref{eq:total_full}}
  back-propagate $\mathcal{L}$; clip gradient norm to $1.0$; AdamW step on $(z,\rho)$\;
}
$a^\star \leftarrow 0.85+0.30\,\operatorname{sigmoid}(\rho)$;\quad $R^\star \leftarrow \operatorname{norm}(\mathcal{B}(\mathcal{U}(z)))\odot G$\;
$\delta^\star \leftarrow \operatorname{clip}\big(\varepsilon\tanh(a^\star P + rR^\star)\odot G,\,-\varepsilon,\varepsilon\big)$\;
\Return{$\delta^\star$}\;
\caption{Supplementary optimization details}
\label{alg:airflow_full}
\end{algorithm}

\textbf{Optimization.} The only trainable parameters are the residual latent
$z\in\mathbb{R}^{d\times d}$ and the amplitude logit $\rho$ (the prior $P$ and
gate $G$ are fixed). We optimize them with AdamW (default betas and
$\epsilon_{\mathrm{Adam}}$, weight decay $10^{-4}$) at learning rate
$\eta=0.095$ for $T=800$ steps, with 30 warmup steps followed by cosine decay
and gradient-norm clipping at $1.0$; the perturbation is clipped to the
$L_\infty$ ball via the $\tanh$ and $\operatorname{clip}$ in \eqref{eq:gen_out}
at every step, so $\lVert\delta\rVert_\infty\le\varepsilon$ holds by
construction. Because $z$ and $\rho$ are input-agnostic and the batch gradient
is averaged over images, the result is a single \emph{universal} perturbation.
At convergence we freeze $(z^\star,\rho^\star)$ and read off the deployable
$\delta^\star$ from \eqref{eq:gen_out}; at attack time it is added to any IR
input with no further optimization and no access to the target model, which is
what makes the threat model gray-box.

\clearpage
\section{Extended Discussion}
\label{app:discussion}

\textbf{Why the attack transfers and which models are most exposed.} The strong cross-model transfer across architectures with different pretraining data and encoders suggests that IR-trained models converge to similar representations of thermal patterns; the airflow perturbation exploits low-level, spatially correlated thermal-texture features that underlie IR scene understanding across models, analogous to how high-frequency patterns transfer across RGB models~\cite{ilyas2019} but with physical grounding that targets universally learned features. Notably, among the non-surrogate backbones the two remote-sensing-specialized models (RemoteCLIP-B32, GeoRSCLIP-B32) are the most vulnerable (50.6\% and 51.7\% ASR, second only to the 54.2\% surrogate), suggesting that domain-specific training enriches the very thermal-phenomenon representations the attack exploits. Among VLMs, the effect is concentrated on global scene identity---which degrades significantly---while localized object presence is largely robust, indicating the perturbation corrupts the global scene representation rather than local object evidence.

\textbf{The IR-cue paradox.} On two of six VLMs (most strikingly Qwen2.5-VL-7B, 86.59$\rightarrow$94.43), IR-cue detection accuracy \emph{rises} under attack: the structured airflow pattern is interpreted as genuine thermal evidence (temperature gradients, convection signatures), making the model \emph{more} confident in an incorrect analysis. This failure mode has no direct analog in RGB attacks and underscores the need for modality-specific robustness evaluation, though its model-dependence indicates it is not universal.

\textbf{Limitations.} Our attack is evaluated in the digital domain; physical realizability through controlled heat sources remains to be demonstrated, though the correlation with natural airflow patterns (0.893) suggests feasibility. We do not evaluate defenses, and our comparison targets IR-specific physical attacks rather than unstructured digital perturbations. Finally, the transfer split may share caption vocabulary with the optimization set, and whether the airflow prior's higher plausibility confers a practical advantage (e.g., evading detectors or surviving input transformations) remains open---all directions we leave for future work.

\clearpage
\section{LLM-as-Judge Evaluation}
\label{app:llmjudge}

We additionally evaluate caption and VQA degradation with a multimodal
LLM-as-Judge protocol. The judge is gpt-4o-mini through ChatAnywhere and receives
the image, the clean answer, and the adversarial answer under the same task
prompt. For each item, the raw attack score is the drop from the clean answer
score to the adversarial answer score after clamping negative drops to zero.
Because our goal is a subtle thermal-airflow disturbance rather than a visibly
destructive artifact, we also report a perturbation-normalized efficiency score,
\(\mathrm{Attack}/\mathrm{MAE}\), computed as the common-clean attack score
divided by the image mean absolute error. The raw score alone favors some
stronger visible baselines, while the normalized score measures attack effect
per unit image distortion and is therefore the metric most aligned with our
setting.

\begin{table}[H]
  \caption{Overall LLM-as-Judge comparison on Qwen2.5-VL cases. Raw attack
  measures answer degradation; Attack/MAE measures degradation per unit visible
  perturbation. Higher is stronger for Raw attack and Attack/MAE, while lower
  MAE and higher SSIM indicate better visual preservation.}
  \label{tab:llm_judge_overall}
  \centering
  \begin{tabular*}{\textwidth}{@{\extracolsep{\fill}}ccccc@{}}
    \toprule
    Method & Raw attack & MAE & SSIM & Attack/MAE \\
    \midrule
    Ours & 18.000 & 0.017 & 0.891 & \textbf{1073.489} \\
    Thermal turbulence & \textbf{30.104} & 0.042 & 0.804 & 718.355 \\
    IR stripe & 21.375 & 0.075 & 0.714 & 284.575 \\
    Hot/cold block & 23.209 & 0.085 & 0.748 & 273.726 \\
    Thermal drift & 16.854 & 0.075 & 0.788 & 223.825 \\
    \bottomrule
  \end{tabular*}
\end{table}

\vspace{-3.0em}

\begin{table}[H]
  \caption{Task-wise LLM-as-Judge efficiency. Values are Attack/MAE; higher
  means stronger answer degradation per unit image distortion.}
  \label{tab:llm_judge_task_efficiency}
  \centering
  \begin{tabular*}{\textwidth}{@{\extracolsep{\fill}}ccccc@{}}
    \toprule
    Method & Caption & Scene & Objects & IR cues \\
    \midrule
    Ours & \textbf{730.569} & \textbf{1103.308} & \textbf{1734.460} & \textbf{725.619} \\
    Thermal turbulence & 584.630 & 640.301 & 930.635 & 717.854 \\
    IR stripe & 187.493 & 235.208 & 519.224 & 196.373 \\
    Hot/cold block & 243.751 & 286.988 & 341.053 & 223.111 \\
    Thermal drift & 66.400 & 323.143 & 314.299 & 191.458 \\
    \bottomrule
  \end{tabular*}
\end{table}

\clearpage
\section{Reproducibility Details and Compute Budget}
\label{app:repro}

This appendix records the practical details needed to reproduce our results:
the software and hardware environment and its compute budget
(Appendix~\ref{app:env}), data preparation and model checkpoints
(Appendix~\ref{app:data}), the exact implementation and hyperparameter
configuration (Appendix~\ref{app:impl}), the standard evaluation and statistical
protocol (Appendix~\ref{app:eval}), and the LLM-as-Judge evaluation
(Appendix~\ref{app:llmjudge}). All values below are the reference
configuration used throughout the main paper unless a per-experiment override is
stated. Some wall-clock values are approximate because the raw GPU accounting
logs were not retained for every exploratory run; when exact timestamps are
unavailable, we report conservative estimates derived from the executed scripts
and run logs.

\subsection{Compute Environment and GPU Budget}
\label{app:env}

The entire pipeline---surrogate fine-tuning, perturbation optimization, and all
evaluations---runs on NVIDIA RTX 4090 (24\,GB) GPUs. The compute footprint is
modest by design: the attack is universal, so one perturbation is optimized once
per surrogate and then applied without any per-image or per-target optimization,
and the only trainable parameters are the $32{\times}32$ residual latent and a
scalar amplitude (Appendix~\ref{app:generator}), 325 values in total.
Peak GPU memory during perturbation optimization stays below 22\,GB, so no
gradient checkpointing or model sharding is required, and a single 24\,GB card
is sufficient. To parallelize the five-backbone study, we place one CLIP model
per GPU (five cards, indices $0$--$4$) and optimize/evaluate them concurrently;
this is a throughput convenience, not a requirement---each run is
single-GPU. The software stack is PyTorch~2.5.1 with CUDA~12.4
on Python~3.12.3; the CLIP backbones and VLMs are loaded from the fixed infrared-domain
model package listed in Appendix~\ref{app:data}.

Table~\ref{tab:budget} breaks down the wall-clock cost of each stage. The
dominant total cost is VLM inference, because six generative models are queried
over four tasks. The dominant train-time attack cost is optimizing the universal
perturbation for $T=800$ steps; because the result is input-agnostic, this cost
is amortized over the entire test set and never repeated at attack time. VLM
evaluation is inference-only (no gradients), and its cost is set by the number
of generated tokens per task rather than by the attack. The total end-to-end
budget to reproduce every number in the paper is approximately 60\,GPU-hours.

\begin{table}[t]
  \caption{Compute budget on RTX 4090 (24\,GB) GPUs. Perturbation
  optimization is a one-time train-time cost amortized over all downstream
  evaluations; inference stages carry no optimization. Wall-clock values are
  approximate when complete scheduler logs were unavailable.}
  \label{tab:budget}
  \centering
  \resizebox{\textwidth}{!}{%
  \begin{tabular}{@{}ccccc@{}}
    \toprule
    Stage & Type & Runs & Wall-clock & GPU-hours \\
    \midrule
    Surrogate IR fine-tuning (OpenAI-CLIP-B32) & train & 1 & $\approx$25\,min & $\approx$0.4 \\
    UAP optimization ($T{=}800$ steps) & train & 1 & 4--6\,min & $\approx$0.1 \\
    CLIP zero-shot eval (5 backbones, 9{,}720 samples) & inference & 5 & $\approx$15\,min & $\approx$0.5 \\
    Cross-model transfer probe (5 backbones, 1{,}000-pool) & inference & 5 & $\approx$5\,min & $\approx$0.1 \\
    VLM eval (6 models, 4 tasks, 1{,}000 samples) & inference & 6 & $\approx$10--12\,h & $\approx$50 \\
    Ablations (loss/position/hyperparameter sweeps) & train+infer & 94 & $\approx$2\,h & $\approx$8 \\
    \midrule
    \textbf{Total} & & & & $\approx$60 \\
    \bottomrule
  \end{tabular}%
  }
\end{table}

\subsection{Dataset Construction and Model Checkpoints}
\label{app:data}

\textbf{Infrared-sample filtering.} The evaluation corpus is assembled from five public
remote-sensing caption datasets: NWPU-Caption~\cite{nwpucaption},
RSICD~\cite{rsicd}, RSITMD~\cite{rsitmd}, RS5M~\cite{rs5m}, and
SkyScript~\cite{skyscript}. All are filtered to infrared samples so that no
RGB information leaks into the thermal evaluation. A sample is retained only if
its filename or metadata carries an explicit IR indicator and carries no RGB
indicator; ambiguous samples are discarded rather than guessed. After filtering,
the test split contains 10{,}000 IR image--text pairs, of which 9{,}720 carry a
remote-sensing scene-category label and form the CLIP zero-shot classification
set; the validation split used for all ablations contains 416 samples. The VLM
study uses a separate 1{,}000-image diagnostic pool annotated for the four
downstream tasks. This diagnostic pool is not used for surrogate fine-tuning,
perturbation optimization, or hyperparameter selection.

\textbf{Checkpoints.} All target models are evaluated from fixed infrared-domain
checkpoints; during attack optimization, only the surrogate OpenAI-CLIP-B32
checkpoint is accessed. The exact checkpoint path for each of the five CLIP
backbones and six VLMs is recorded in the model-package manifest
file, which pins the reported numbers to a fixed set of weights.

\subsection{Implementation Details}
\label{app:impl}

Table~\ref{tab:hparams} consolidates every hyperparameter needed to reproduce
the reference configuration; all values match the main paper and the equations
of Appendix~\ref{app:method}. The perturbation is the prior-plus-residual
parameterization of Appendix~\ref{app:generator} (latent $d{=}32$, residual
scale $r{=}0.60$); the airflow prior of Appendix~\ref{app:templates} is loaded
once and held fixed throughout optimization. Unless noted, each ablation sweeps
a single hyperparameter with all others held at these reference values.

\begin{table}[t]
  \caption{Reference hyperparameters (the $\varepsilon{=}100$, full-position
  configuration). All ablations vary one row at a time with the rest held fixed.
  The IR input resolution follows each backbone's native CLIP preprocessing.}
  \label{tab:hparams}
  \centering
  \begin{tabular}{@{}ccc@{}}
    \toprule
    Symbol & Meaning & Value \\
    \midrule
    $\varepsilon$ & $L_\infty$ budget (of 255) & 100 \\
    $T$ & optimization steps & 800 \\
    $\eta$ & learning rate (AdamW, wd $10^{-4}$) & 0.095 \\
    -- & warmup / schedule & 30 steps / cosine \\
    -- & gradient-norm clip & 1.0 \\
    $d$ & residual latent size ($d{\times}d$) & 32 \\
    $r$ & residual scale & 0.60 \\
    -- & residual blur kernel / $\sigma$ & 9 / 2.5 \\
    $\gamma$ & gate threshold & 0.01 \\
    $\mu$ & mean-magnitude budget & 0.16 \\
    $w_{\text{conf}}$ & confidence-ratio weight & 8.0 \\
    $w_{\text{air}}$ & airflow prior weight & 1.0 \\
    -- & optimizer & AdamW (betas/epsilon default) \\
    -- & batch size (L14 / others) & 64 / 256 \\
    -- & IR input resolution $H{\times}W$ & native (e.g.\ $224{\times}224$) \\
    -- & random seed & fixed \\
    \bottomrule
  \end{tabular}
\end{table}

\subsection{Evaluation Protocol}
\label{app:eval}

\textbf{CLIP.} Each backbone acts as a zero-shot scene classifier: an IR image
is assigned the scene category whose text prompt maximizes image--text cosine
similarity. ASR is the paired top-1 flip rate---the fraction of samples whose
adversarial top-1 class differs from the clean top-1 class on the same
image. This is a prediction-stability ASR: it measures whether the attack
changes the model's own clean decision, and it does not assume that the clean
top-1 class is always the human ground-truth label. For labeled scene
experiments, the evaluation logs retain clean top-1, adversarial top-1, and the
annotated label so that clean accuracy and robust accuracy can be audited
separately from flip ASR. The cross-model transfer study of the main paper
instead applies one UAP to held-out samples and
measures nearest-caption retrieval flips over a 1{,}000-sample candidate pool,
which is a strictly more sensitive probe and therefore reports higher numbers
that are not comparable to ASR.

\textbf{VLMs.} We report ROUGE-L for captioning, scene accuracy, object F1, and
IR-cue accuracy, each on the full 1{,}000-sample set. Because clean and attacked
outputs are paired on the same images, binary scene and IR-cue metrics are
reported with Wilson confidence intervals and paired tests when per-sample
correctness labels are available. ROUGE-L and object F1 are continuous or
set-valued scores, so we treat their statistical comparisons as descriptive
unless a paired bootstrap or permutation test is explicitly reported. The main
claims are therefore based on effect sizes that are consistent across models,
not on unpaired significance tests. The exact task prompts issued to each VLM
are defined verbatim in the evaluation script vlm\_eval\_generate.py
so that generation conditions can be reproduced.

\subsection{Reproducibility Statement}
\label{app:reprostmt}

The appendix specifies the full method (Appendix~\ref{app:method}), every
reference hyperparameter (Table~\ref{tab:hparams}), the data preparation
procedure (Appendix~\ref{app:data}), the standard evaluation and statistical
protocol (Appendix~\ref{app:eval}), and the LLM-as-Judge protocol
(Appendix~\ref{app:llmjudge}), which together determine all reported numbers.
The implementation is organized around explicit split files, model
manifests, perturbation checkpoints, and evaluation scripts, so the experiment
can be reproduced from the public source datasets and the fixed infrared-domain
model package described above. All attack optimization runs are single-GPU jobs;
multi-GPU execution is used only to run independent models in parallel.

\clearpage
\section{Additional Qualitative Diagnostics}
\label{app:qualitative}

The quantitative results in the main paper measure whether AirflowAttack changes
model predictions, but they do not by themselves show whether the resulting
images remain meaningful infrared remote-sensing samples. We therefore perform a
paired qualitative audit on clean images, perturbation-only visualizations,
adversarial images, CLIP prediction changes, VLM outputs, and attention maps. The
purpose of this audit is not to introduce a new metric, but to verify that the
reported ASR is obtained by a structured thermal-airflow pattern rather than by
destroying the scene content.

\subsection{Perturbation Visibility}
\label{app:qual_visibility}

The main experiments use $\varepsilon=100$ on the 8-bit intensity scale. This
number should be interpreted as an $L_\infty$ upper bound rather than as the
typical pixel change: the perturbation is produced by a fixed airflow prior, a
blurred low-dimensional residual, and a soft spatial gate, so its visible energy
is spatially coherent instead of independently saturating all pixels. In the
reference OpenAI-CLIP-B32 run, the selected perturbation has mean absolute
magnitude about $3.4/255$ and peak magnitude about $87.6/255$ after gating,
showing that the nominal budget is used only locally. In the visual examples,
the adversarial image preserves the large-scale scene layout, object boundaries,
and thermal contrast structure of the clean image, while the added pattern
appears as a weak airflow-like texture. This is the reason we use
$\varepsilon=100$ as the reference setting rather than the higher budgets in the
strength ablation: larger budgets further increase attack success, but they also
make the perturbation less plausible as a natural thermal disturbance.

\subsection{Prediction-Level Consistency Checks}
\label{app:qual_clip}

For every qualitative CLIP case, we inspect the clean top-1 class and the
adversarial top-1 class on the same image. This paired design is important
because the clean label may differ from the human semantic label in some
remote-sensing scenes; ASR is therefore counted only as a clean-to-adversarial
top-1 flip, not as disagreement with an external annotation. Successful cases
typically preserve the human-recognizable scene while changing the CLIP decision
to a visually incompatible category. The attention maps in the main paper show
the same mechanism spatially: the clean model concentrates on scene-defining
regions, whereas the adversarial model response is pulled toward the structured
thermal texture, producing the top-1 flip without requiring object removal or
geometric distortion.

\subsection{Caption and VQA Failure Modes}
\label{app:qual_vlm}

The VLM examples reveal a different failure pattern from the CLIP classifier. In
captioning, attacked outputs often become more generic and less tied to the
actual scene, which explains why ROUGE-L only partially reflects the qualitative
degradation: a generic caption can still share common words with the reference
caption. In VQA, the strongest and most consistent effect is on scene identity.
Object-level answers are more stable because localized high-contrast structures
often remain visible after the airflow perturbation. Infrared-cue questions show
a modality-specific failure: some VLMs interpret the artificial airflow texture
as genuine thermal evidence, such as temperature gradients or convection-like
patterns, and therefore answer with greater confidence in an incorrect thermal
interpretation. This behavior supports the claim that IR-specific perturbations
can exploit thermal semantics rather than merely adding generic visual noise.

\subsection{Failure Cases}
\label{app:qual_failures}

The attack is weakest when the clean image contains a dominant, high-contrast
object or a simple scene layout whose semantic evidence is spatially localized.
In these cases, the airflow perturbation may change the model confidence without
changing the top-1 class. It is also less effective on scenes whose clean
prediction margin is large, especially for the larger ViT-L/14 backbone, where a
substantial confidence drop does not always translate into a classification flip.
These failure cases are consistent with the quantitative results: AirflowAttack
primarily disrupts global thermal texture and scene-level representations, while
localized object evidence is comparatively more robust. We therefore frame the
method as a universal scene-understanding attack for infrared remote sensing, not
as a guaranteed object-removal or object-detection attack.

\clearpage
\section{Dataset Statistics and Sampling Details}
\label{app:dataset_stats}

This appendix gives the concrete split statistics used by the experiments. The
infrared corpus is built from NWPU-Caption, RSICD, RSITMD, RS5M, and SkyScript.
Each retained sample has an infrared-style image and an IR-aware caption; samples
with RGB indicators are filtered out before any training or evaluation. The
surrogate fine-tuning split contains 48{,}616 image--caption pairs, the ablation
validation split contains 416 pairs, and the held-out test split contains
10{,}000 pairs. For the VLM transfer study and the short-class diagnostic
classifier outputs, we additionally use a 1{,}000-image balanced diagnostic
subset with explicit source and scene labels. This subset is a held-out
diagnostic pool: it is used for transfer evaluation and qualitative analysis,
not for surrogate training, loss selection, or hyperparameter tuning.

The held-out test split is intentionally RS5M-heavy because it follows the
available infrared sample pool after filtering. The diagnostic subset is more
balanced across sources and is therefore used for qualitative inspection,
downstream VLM transfer, and source-wise analysis. Its labels cover 169 unique
scene names; the most frequent labels are bridge, airplane, airport, river,
farmland, parking, industrial, harbor, tennis, and resort/beach, each appearing
between 21 and 34 times. We use this diagnostic subset only for secondary
analysis; the main CLIP ASR comparison and VLM tables remain the primary
reported results.

\begin{table}[h]
  \caption{Infrared split statistics by source dataset. The diagnostic 1{,}000
  subset is the labeled subset used for VLM transfer inputs and the per-source
  diagnostics in Appendix~\ref{app:source_diag}.}
  \label{tab:dataset_splits}
  \centering
  \resizebox{0.8\textwidth}{!}{%
  \begin{tabular}{@{}ccccccc@{}}
    \toprule
    Split & NWPU & RSICD & RSITMD & RS5M & SkyScript & Total \\
    \midrule
    Surrogate train & 2{,}190 & 711 & 276 & 40{,}580 & 4{,}859 & 48{,}616 \\
    Ablation validation & 47 & 0 & 0 & 369 & 0 & 416 \\
    Held-out test & 515 & 164 & 87 & 8{,}138 & 1{,}096 & 10{,}000 \\
    Diagnostic/VLM subset & 205 & 166 & 83 & 332 & 214 & 1{,}000 \\
    \bottomrule
  \end{tabular}%
  }
\end{table}

\clearpage
\section{Baseline Implementation Details}
\label{app:baseline_details}

All physical baselines are implemented as image-space infrared perturbations and
are applied once to each clean IR image, with no gradient access to the target
model and no per-image target optimization. Random fields are generated with the
same fixed seed used by our method. The perturbation is always
clipped to the same valid image range as AirflowAttack, and the final
adversarial image is clipped to $[0,1]$. For the CLIP comparison table, all
methods are constrained by the same nominal $L_\infty$ budget, so the reported
ASR values are directly comparable.

\textbf{Atmospheric thermal turbulence.} The turbulence baseline uses a
low-frequency two-channel displacement field and a separate low-frequency
scintillation field. The displacement field is decoded from a $14{\times}14$
latent map and scaled to a maximum displacement of approximately 5.5 pixels
before bilinear resampling with reflection padding. The scintillation field is
decoded from a $10{\times}10$ latent map and applied multiplicatively with
strength 0.065. A Gaussian blur with kernel size 5 and $\sigma=1.0$ is then
applied to emulate the local smoothness of atmospheric distortion.

\textbf{IR fixed-pattern stripe noise.} The stripe baseline simulates sensor
readout artifacts by combining a column-wise low-frequency field and a weaker
row-wise field. The column component receives weight 0.75 and the row component
weight 0.25; the combined pattern is peak-normalized and added with amplitude
$0.85\varepsilon$. This produces coherent vertical banding rather than
independent pixel noise.

\textbf{Thermal hot/cold blocks.} The hot/cold baseline overlays seven smooth
signed thermal blobs. Each blob has a random center in the central image region,
random horizontal and vertical scales in the range 0.045--0.115 of the image
width/height, and a random hot or cold sign. The summed pattern is blurred with
a Gaussian kernel of size 11 and $\sigma=2.5$, mean-removed, peak-normalized,
and applied with amplitude $0.95\varepsilon$.

\textbf{Thermal sensor drift/non-uniformity.} The sensor-drift baseline decodes
a smooth $8{\times}8$ low-frequency field and uses it as both multiplicative
gain and additive offset. Specifically, the clean image is transformed as
$x\cdot(1+0.18F)+0.35\varepsilon F$, where $F$ is the normalized drift field.
This approximates non-uniform detector gain, offset drift, and slow thermal
calibration error.

\clearpage
\section{Qualitative Case Gallery}
\label{app:case_gallery}

Following the table-style appendix layout used in the reference paper, we give
additional qualitative cases in which each row pairs the clean image, the
adversarial image, the clean model output, and the attacked output. The goal is
not to introduce new metrics, but to make the paired failure mode visible: the
scene remains recognizable to a human observer, while the model output shifts
toward a different scene or a more generic thermal description.

\begin{table}[H]
  \caption{Qualitative VLM case group 1. Each row shows the clean input,
  adversarial input, and concise paired model outputs.}
  \label{tab:qual_vlm_case_table_group1}
  \centering
  \scriptsize
  \setlength{\tabcolsep}{2pt}
  \resizebox{\textwidth}{!}{%
  \begin{tabular}{@{}M{1.45cm}C{1.62cm}C{1.62cm}M{3.6cm}M{3.6cm}M{2.1cm}@{}}
    \toprule
    Case & Clean image & Adv. image & Clean output & Adv. output & Shift type \\
    \midrule
    case\_01 NWPU &
    \includegraphics[width=1.55cm]{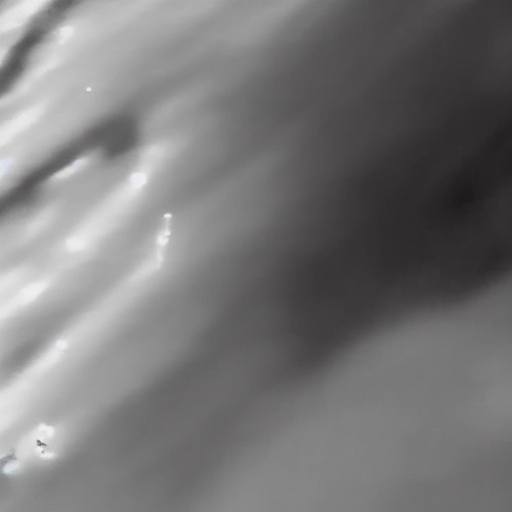} &
    \includegraphics[width=1.55cm]{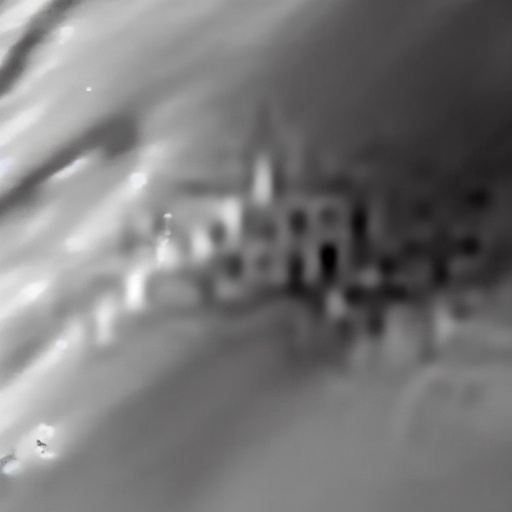} &
    Low-texture bare land with faint structural outlines and scattered trees. &
    High-contrast urban scene with buildings, roads, bright spots, and
    low-texture regions. &
    Bare land to city. \\
    \midrule
    case\_02 SkyScript &
    \includegraphics[width=1.55cm]{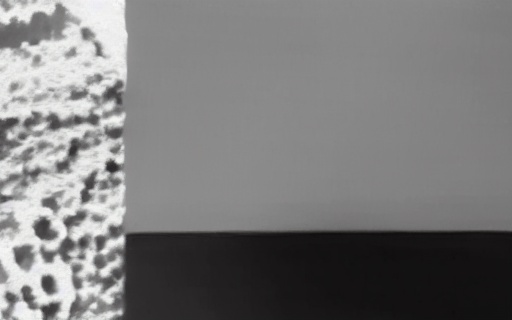} &
    \includegraphics[width=1.55cm]{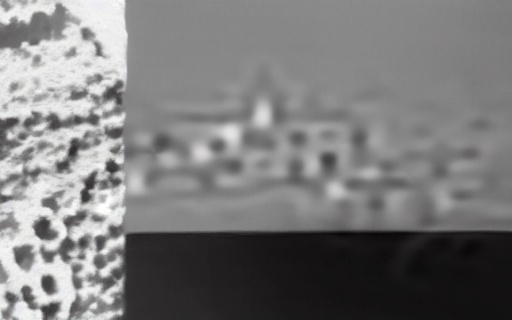} &
    Beach and water separated by a clear shoreline boundary. &
    Urban scene with structural outlines, river-like smooth gradients, and
    buildings. &
    Natural scene to urban. \\
    \midrule
    case\_03 SkyScript &
    \includegraphics[width=1.55cm]{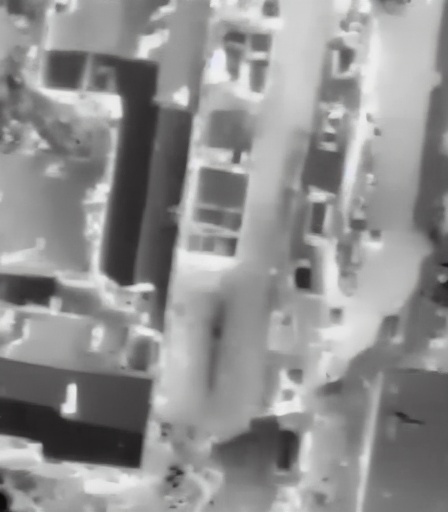} &
    \includegraphics[width=1.55cm]{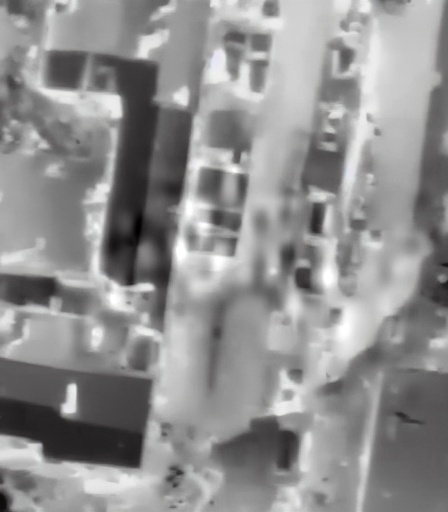} &
    Buildings with distinct shapes adjacent to railway-station-like regions. &
    Residential area with houses, streets, and reflective open spaces. &
    Specific structure to generic residential. \\
    \bottomrule
  \end{tabular}%
  }
\end{table}

\begin{table}[H]
  \caption{Qualitative VLM case group 2.}
  \label{tab:qual_vlm_case_table_group2}
  \centering
  \scriptsize
  \setlength{\tabcolsep}{2pt}
  \resizebox{\textwidth}{!}{%
  \begin{tabular}{@{}M{1.45cm}C{1.62cm}C{1.62cm}M{3.6cm}M{3.6cm}M{2.1cm}@{}}
    \toprule
    Case & Clean image & Adv. image & Clean output & Adv. output & Shift type \\
    \midrule
    case\_04 SkyScript &
    \includegraphics[width=1.55cm]{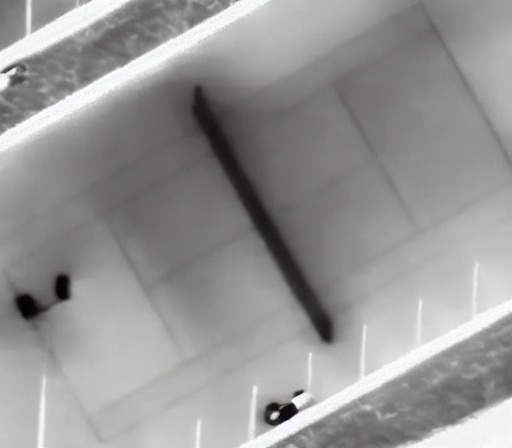} &
    \includegraphics[width=1.55cm]{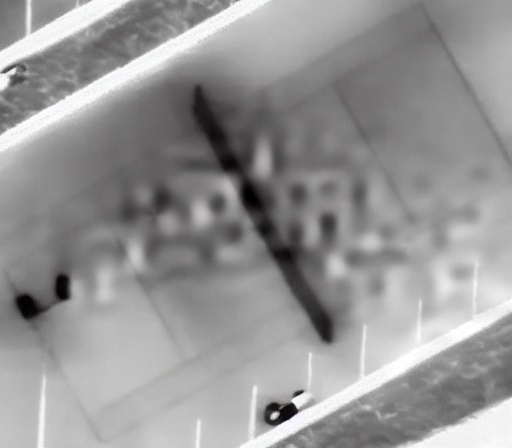} &
    Tennis court with dark boundary lines and a smooth playing surface. &
    Tennis court remains, but the response over-emphasizes thermal intensity
    patterns around it. &
    Cue distortion. \\
    \midrule
    case\_05 NWPU &
    \includegraphics[width=1.55cm]{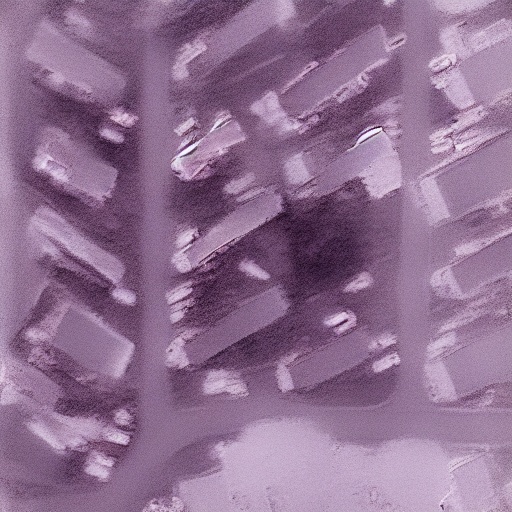} &
    \includegraphics[width=1.55cm]{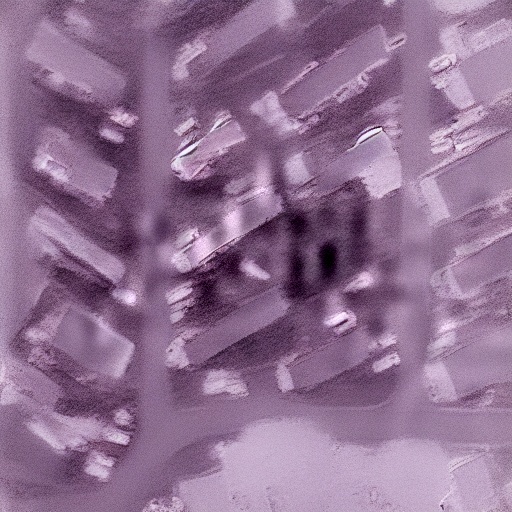} &
    Mobile-home park with roads and surrounding darker tree regions. &
    Residential area with parked cars, roads, buildings, and grid-like layout. &
    Fine label becomes generic. \\
    \midrule
    case\_06 RSICD &
    \includegraphics[width=1.55cm]{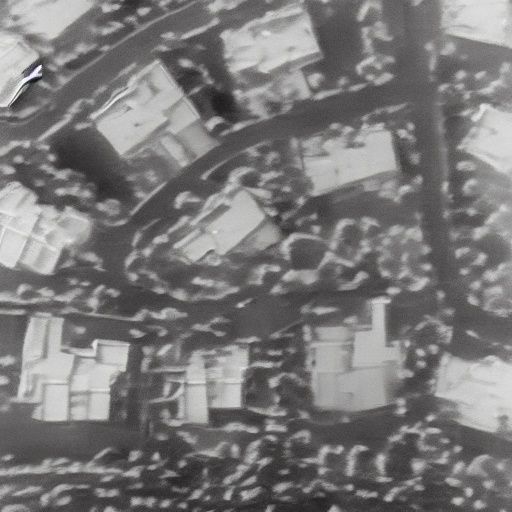} &
    \includegraphics[width=1.55cm]{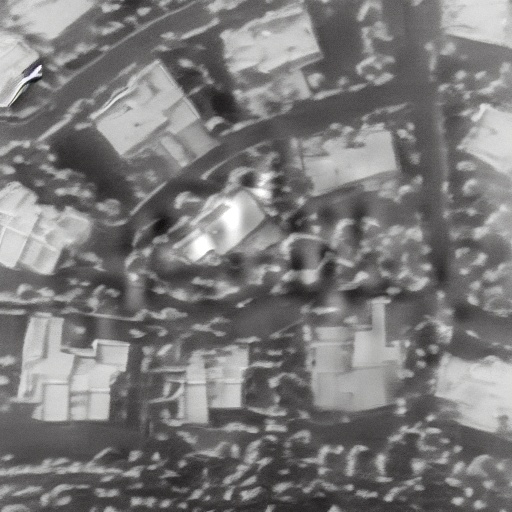} &
    Residential neighborhood with houses, roads, and dense tree cover. &
    Residential area with an added tennis-court-like low-texture region. &
    Object hallucination. \\
    \bottomrule
  \end{tabular}%
  }
\end{table}

\begin{table}[H]
  \caption{Qualitative VLM case group 3.}
  \label{tab:qual_vlm_case_table_group3}
  \centering
  \scriptsize
  \setlength{\tabcolsep}{2pt}
  \resizebox{\textwidth}{!}{%
  \begin{tabular}{@{}M{1.45cm}C{1.62cm}C{1.62cm}M{3.6cm}M{3.6cm}M{2.1cm}@{}}
    \toprule
    Case & Clean image & Adv. image & Clean output & Adv. output & Shift type \\
    \midrule
    case\_07 NWPU &
    \includegraphics[width=1.55cm]{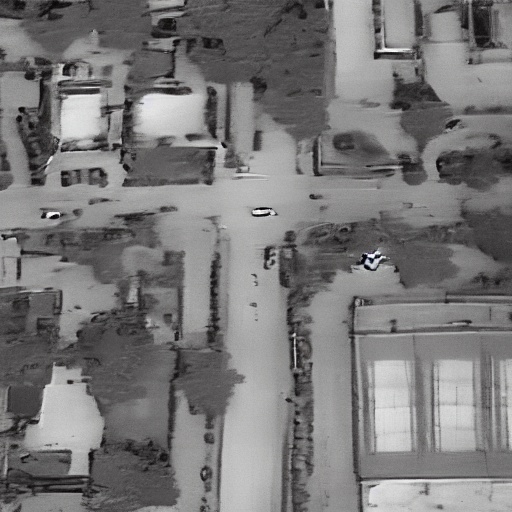} &
    \includegraphics[width=1.55cm]{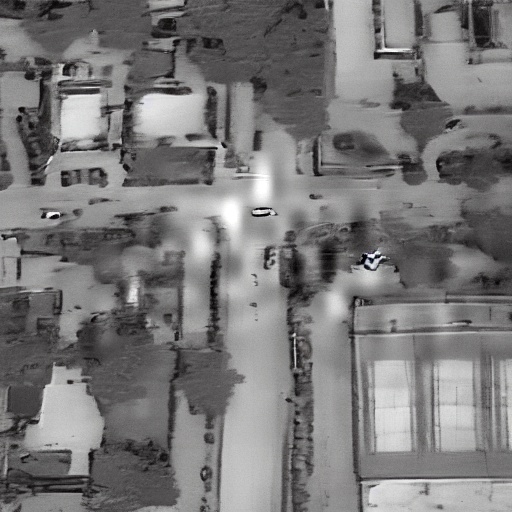} &
    buildings, roads, cars, trees &
    tennis courts, buildings, roads, trees &
    Object answer injects a court cue. \\
    \midrule
    case\_08 RSICD &
    \includegraphics[width=1.55cm]{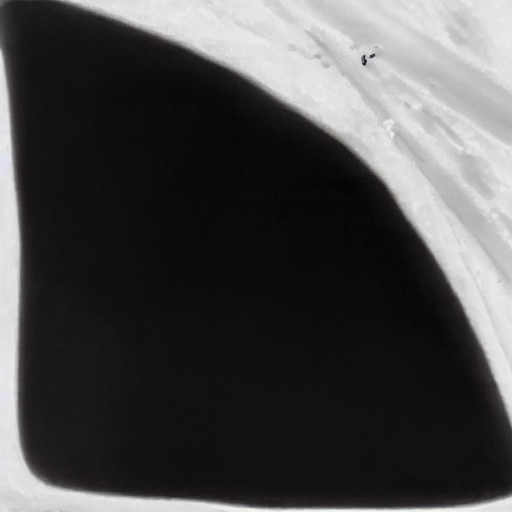} &
    \includegraphics[width=1.55cm]{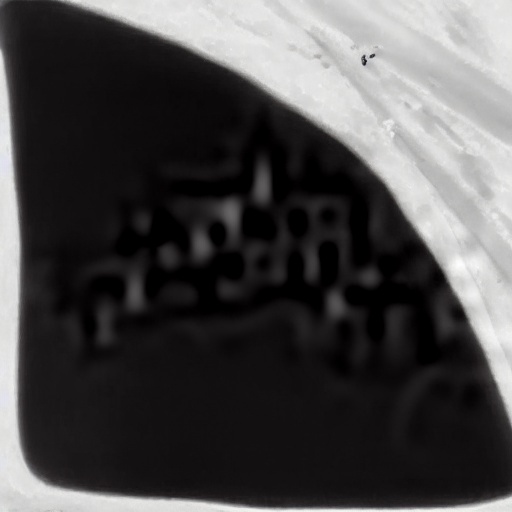} &
    Dark lake scene. &
    City-like aerial scene with buildings, road, and trees. &
    Scene label changes. \\
    \midrule
    case\_09 RSITMD &
    \includegraphics[width=1.55cm]{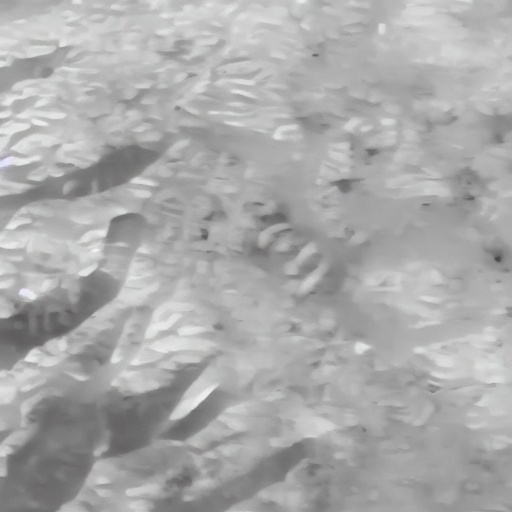} &
    \includegraphics[width=1.55cm]{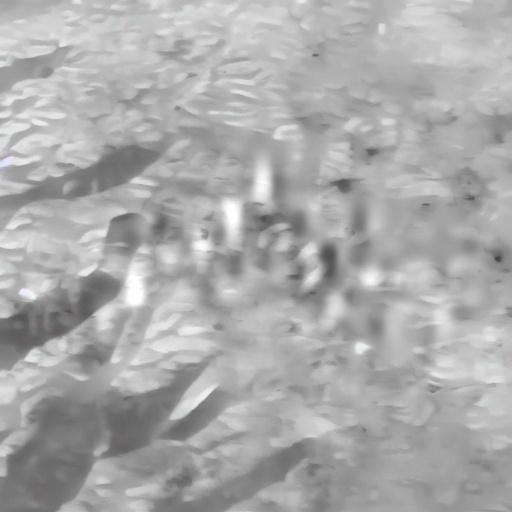} &
    Residential or industrial structures with roads and trees. &
    Urban scene with buildings, roads, trees, and water-body terms. &
    Semantic broadening. \\
    \bottomrule
  \end{tabular}%
  }
\end{table}

\begin{table}[H]
  \caption{Qualitative VLM case group 4.}
  \label{tab:qual_vlm_case_table_group4}
  \centering
  \scriptsize
  \setlength{\tabcolsep}{2pt}
  \resizebox{\textwidth}{!}{%
  \begin{tabular}{@{}M{1.45cm}C{1.62cm}C{1.62cm}M{3.6cm}M{3.6cm}M{2.1cm}@{}}
    \toprule
    Case & Clean image & Adv. image & Clean output & Adv. output & Shift type \\
    \midrule
    case\_10 RSICD &
    \includegraphics[width=1.55cm]{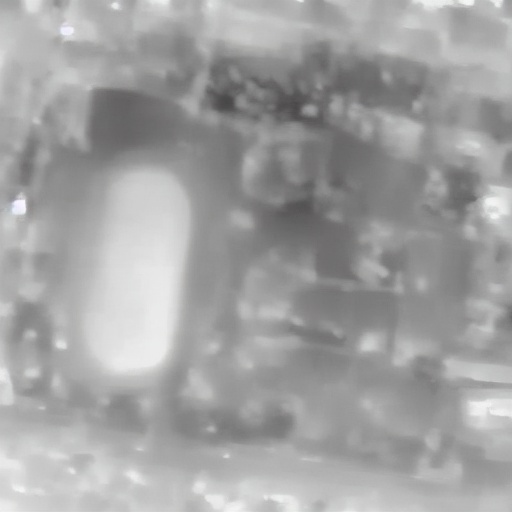} &
    \includegraphics[width=1.55cm]{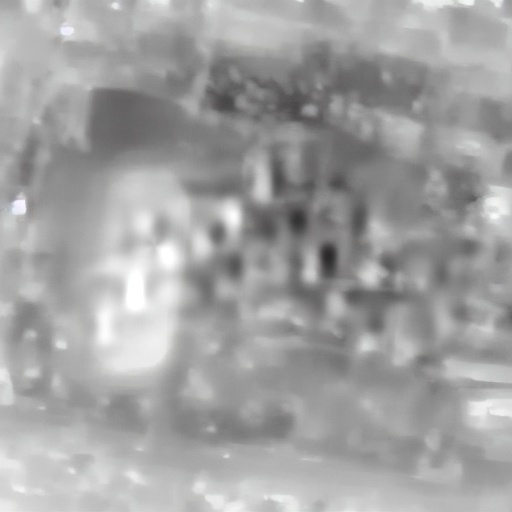} &
    Playground with bright low-texture surface and distinct structural outlines. &
    Residential area with trees, buildings, roads, parking lot, and tennis
    court. &
    Scene and object drift. \\
    \midrule
    case\_11 RS5M &
    \includegraphics[width=1.55cm]{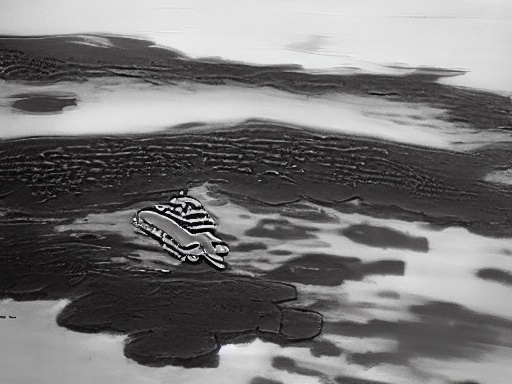} &
    \includegraphics[width=1.55cm]{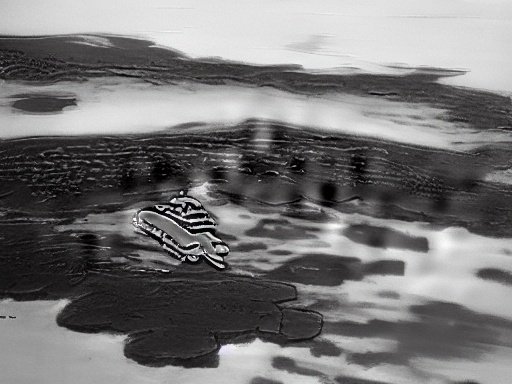} &
    Ocean with varying intensities and a boat as a bright structure. &
    Beach/ocean or land-water contrast; the boat focus is weakened. &
    Key object weakened. \\
    \midrule
    case\_12 RS5M &
    \includegraphics[width=1.55cm]{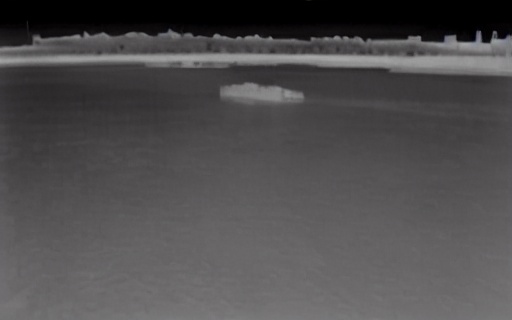} &
    \includegraphics[width=1.55cm]{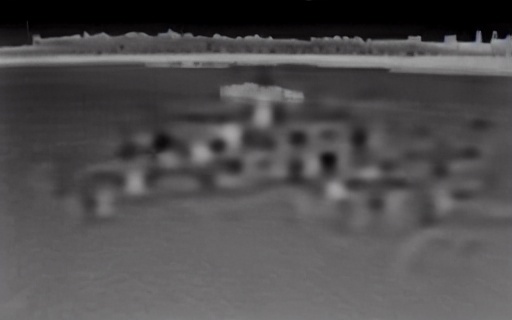} &
    Beach and ocean with smooth low-texture water surface. &
    Dense urban structures and harbor-like high-contrast outlines. &
    Coast to urban harbor. \\
    \bottomrule
  \end{tabular}%
  }
\end{table}

\clearpage
\section{Extended Attention-Map Visualization}
\label{app:extended_heatmap}

We also follow the reference appendix's CAM-style visualization protocol and
include expanded attention examples in table form. For each clean/adversarial
pair, the table lists the model's top-1 scene prediction and shows both the
original image and the layer-12 attention map. The comparison should be read as
a paired change in model grounding, not as a human segmentation mask.

\begin{table}[H]
  \caption{Extended layer-12 attention-map case group 1. Each row includes
  clean and adversarial raw images together with their attention maps.}
  \label{tab:extended_heatmap_table_group1}
  \centering
  \scriptsize
  \setlength{\tabcolsep}{2pt}
  \resizebox{\textwidth}{!}{%
  \begin{tabular}{@{}M{1.25cm}C{1.45cm}C{1.45cm}C{1.45cm}C{1.45cm}M{2.0cm}M{2.0cm}M{2.5cm}@{}}
    \toprule
    Case & Clean & Adv. & Clean heatmap & Adv. heatmap & Clean top-1 & Adv. top-1 & Attention behavior \\
    \midrule
    0538 RSICD &
    \includegraphics[width=1.35cm]{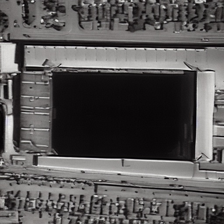} &
    \includegraphics[width=1.35cm]{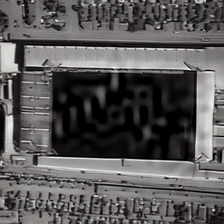} &
    \includegraphics[width=1.35cm]{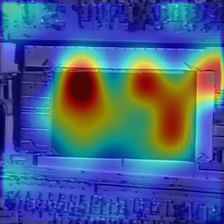} &
    \includegraphics[width=1.35cm]{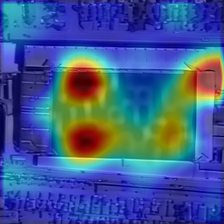} &
    stadium & parking lots &
    Field evidence weakens and shifts to parking-like texture. \\
    \midrule
    0504 RS5M &
    \includegraphics[width=1.35cm]{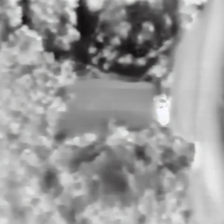} &
    \includegraphics[width=1.35cm]{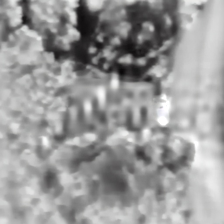} &
    \includegraphics[width=1.35cm]{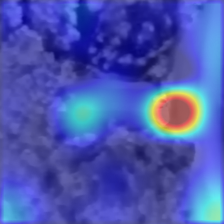} &
    \includegraphics[width=1.35cm]{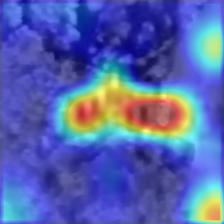} &
    park & church &
    Diffuse land evidence is redirected to compact bright regions. \\
    \midrule
    0507 RS5M &
    \includegraphics[width=1.35cm]{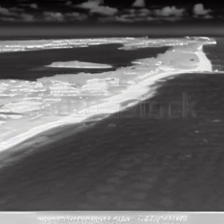} &
    \includegraphics[width=1.35cm]{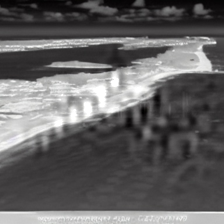} &
    \includegraphics[width=1.35cm]{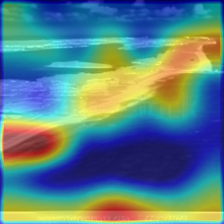} &
    \includegraphics[width=1.35cm]{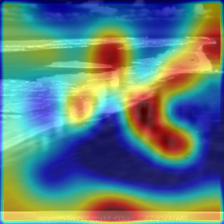} &
    beach & beach &
    Label stays stable, but confidence and grounding are suppressed. \\
    \midrule
    0516 RSICD &
    \includegraphics[width=1.35cm]{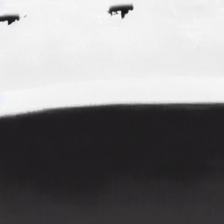} &
    \includegraphics[width=1.35cm]{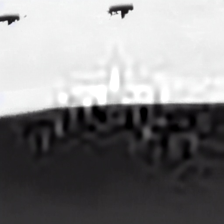} &
    \includegraphics[width=1.35cm]{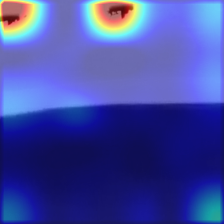} &
    \includegraphics[width=1.35cm]{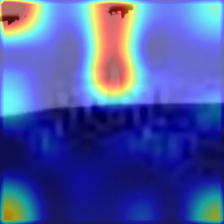} &
    rectangular farmland & tall buildings &
    Low-frequency field evidence becomes building-like. \\
    \bottomrule
  \end{tabular}%
  }
\end{table}

\begin{table}[H]
  \caption{Extended layer-12 attention-map case group 2.}
  \label{tab:extended_heatmap_table_group2}
  \centering
  \scriptsize
  \setlength{\tabcolsep}{2pt}
  \resizebox{\textwidth}{!}{%
  \begin{tabular}{@{}M{1.25cm}C{1.45cm}C{1.45cm}C{1.45cm}C{1.45cm}M{2.0cm}M{2.0cm}M{2.5cm}@{}}
    \toprule
    Case & Clean & Adv. & Clean heatmap & Adv. heatmap & Clean top-1 & Adv. top-1 & Attention behavior \\
    \midrule
    0377 NWPU &
    \includegraphics[width=1.35cm]{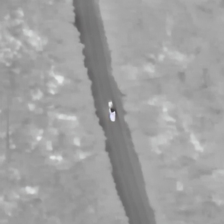} &
    \includegraphics[width=1.35cm]{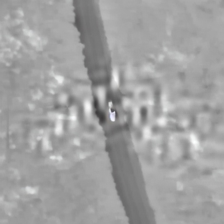} &
    \includegraphics[width=1.35cm]{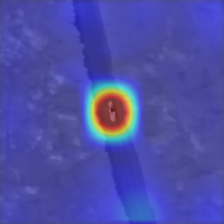} &
    \includegraphics[width=1.35cm]{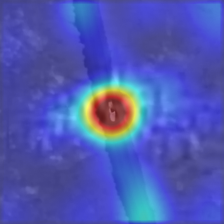} &
    white road & white road &
    Road geometry remains, while confidence is reduced. \\
    \midrule
    0700 RS5M &
    \includegraphics[width=1.35cm]{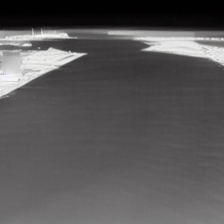} &
    \includegraphics[width=1.35cm]{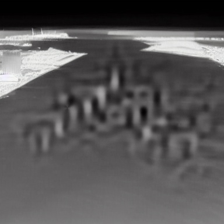} &
    \includegraphics[width=1.35cm]{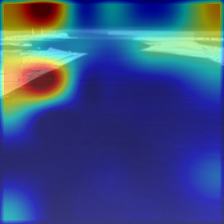} &
    \includegraphics[width=1.35cm]{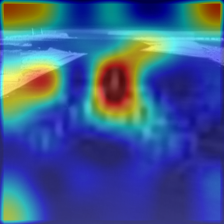} &
    beach & tall buildings &
    Water/shore evidence is redirected to compact structures. \\
    \midrule
    0412 RSICD &
    \includegraphics[width=1.35cm]{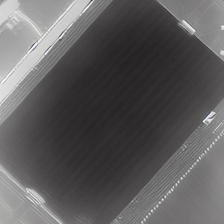} &
    \includegraphics[width=1.35cm]{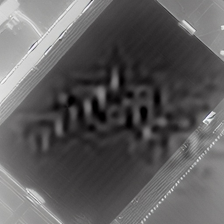} &
    \includegraphics[width=1.35cm]{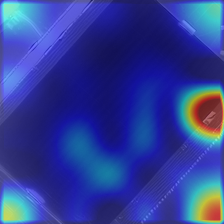} &
    \includegraphics[width=1.35cm]{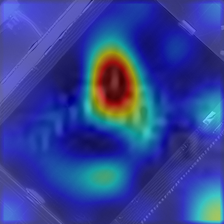} &
    white roofs & church &
    Roof-like evidence is pulled toward object-like responses. \\
    \midrule
    0325 SkyScript &
    \includegraphics[width=1.35cm]{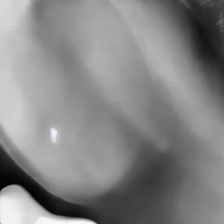} &
    \includegraphics[width=1.35cm]{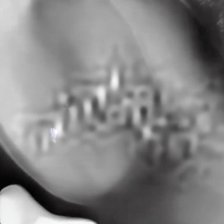} &
    \includegraphics[width=1.35cm]{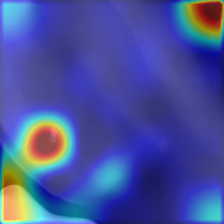} &
    \includegraphics[width=1.35cm]{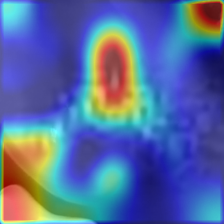} &
    golf course & tall buildings &
    Smooth course texture is replaced by building-like evidence. \\
    \bottomrule
  \end{tabular}%
  }
\end{table}

\begin{table}[H]
  \caption{Extended layer-12 attention-map case group 3.}
  \label{tab:extended_heatmap_table_group3}
  \centering
  \scriptsize
  \setlength{\tabcolsep}{2pt}
  \resizebox{\textwidth}{!}{%
  \begin{tabular}{@{}M{1.25cm}C{1.45cm}C{1.45cm}C{1.45cm}C{1.45cm}M{2.0cm}M{2.0cm}M{2.5cm}@{}}
    \toprule
    Case & Clean & Adv. & Clean heatmap & Adv. heatmap & Clean top-1 & Adv. top-1 & Attention behavior \\
    \midrule
    0020 RS5M &
    \includegraphics[width=1.35cm]{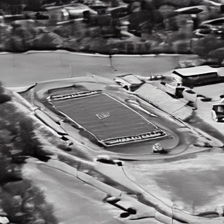} &
    \includegraphics[width=1.35cm]{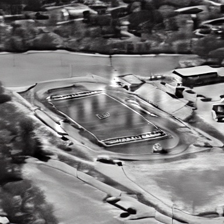} &
    \includegraphics[width=1.35cm]{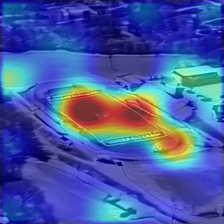} &
    \includegraphics[width=1.35cm]{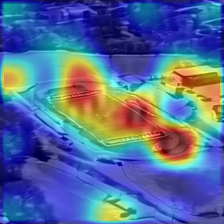} &
    stadium & ground track field &
    Sports-field attention moves to a neighboring field category. \\
    \midrule
    0281 RSICD &
    \includegraphics[width=1.35cm]{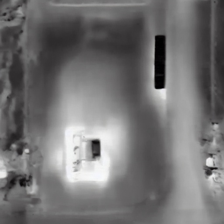} &
    \includegraphics[width=1.35cm]{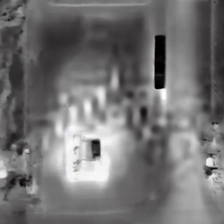} &
    \includegraphics[width=1.35cm]{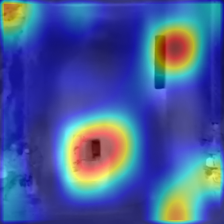} &
    \includegraphics[width=1.35cm]{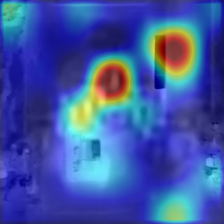} &
    baseball diamond & tall buildings &
    Court-like geometry loses dominance to structural texture. \\
    \midrule
    0650 RS5M &
    \includegraphics[width=1.35cm]{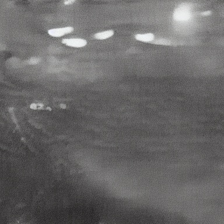} &
    \includegraphics[width=1.35cm]{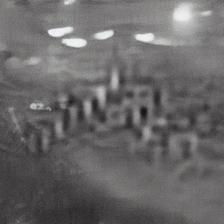} &
    \includegraphics[width=1.35cm]{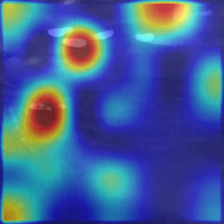} &
    \includegraphics[width=1.35cm]{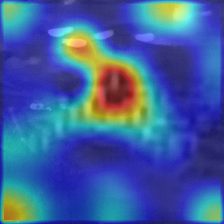} &
    pond & tall buildings &
    Low-texture water evidence changes to clustered structures. \\
    \midrule
    0953 SkyScript &
    \includegraphics[width=1.35cm]{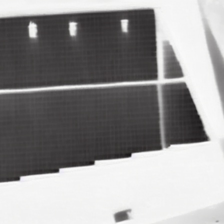} &
    \includegraphics[width=1.35cm]{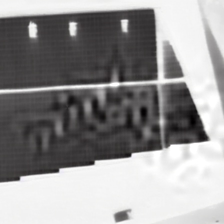} &
    \includegraphics[width=1.35cm]{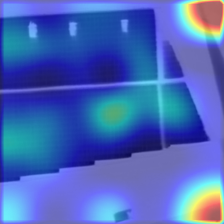} &
    \includegraphics[width=1.35cm]{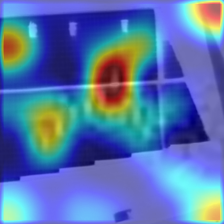} &
    white roofs & white roofs &
    Repeated roof cues stay stable under perturbation. \\
    \bottomrule
  \end{tabular}%
  }
\end{table}

\begin{table}[H]
  \caption{Extended layer-12 attention-map case group 4.}
  \label{tab:extended_heatmap_table_group4}
  \centering
  \scriptsize
  \setlength{\tabcolsep}{2pt}
  \resizebox{\textwidth}{!}{%
  \begin{tabular}{@{}M{1.25cm}C{1.45cm}C{1.45cm}C{1.45cm}C{1.45cm}M{2.0cm}M{2.0cm}M{2.5cm}@{}}
    \toprule
    Case & Clean & Adv. & Clean heatmap & Adv. heatmap & Clean top-1 & Adv. top-1 & Attention behavior \\
    \midrule
    0914 RS5M &
    \includegraphics[width=1.35cm]{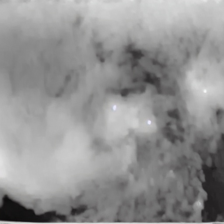} &
    \includegraphics[width=1.35cm]{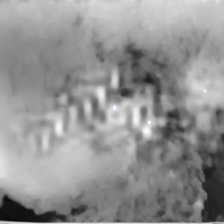} &
    \includegraphics[width=1.35cm]{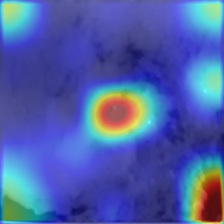} &
    \includegraphics[width=1.35cm]{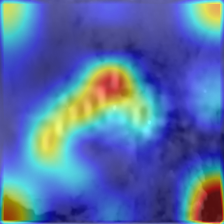} &
    cloud & church &
    Cloud-like texture is pulled toward compact structural evidence. \\
    \midrule
    0227 RS5M &
    \includegraphics[width=1.35cm]{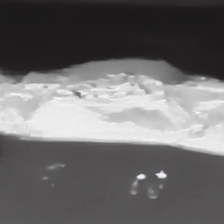} &
    \includegraphics[width=1.35cm]{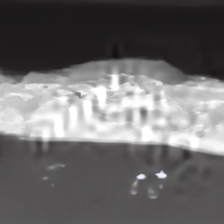} &
    \includegraphics[width=1.35cm]{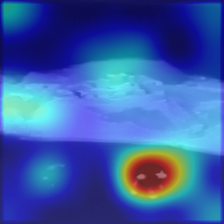} &
    \includegraphics[width=1.35cm]{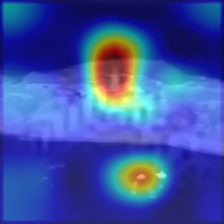} &
    island & harbor &
    Coastline grounding shifts toward harbor-like regions. \\
    \midrule
    0554 RS5M &
    \includegraphics[width=1.35cm]{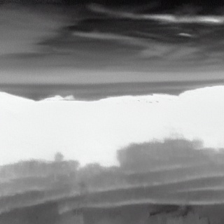} &
    \includegraphics[width=1.35cm]{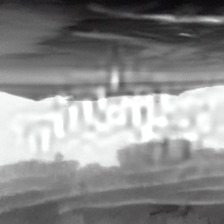} &
    \includegraphics[width=1.35cm]{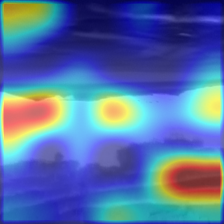} &
    \includegraphics[width=1.35cm]{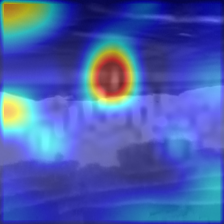} &
    mountain & tall buildings &
    Terrain texture becomes high-response structure. \\
    \midrule
    0829 RS5M &
    \includegraphics[width=1.35cm]{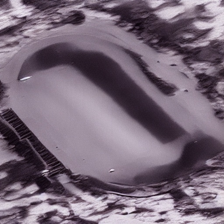} &
    \includegraphics[width=1.35cm]{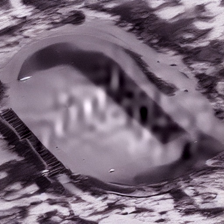} &
    \includegraphics[width=1.35cm]{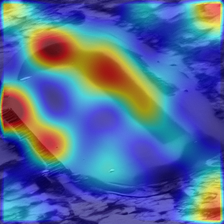} &
    \includegraphics[width=1.35cm]{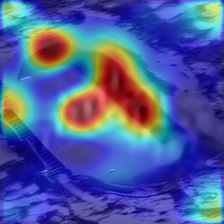} &
    baseball diamond & stadium &
    Field geometry is redirected inside sports-scene semantics. \\
    \bottomrule
  \end{tabular}%
  }
\end{table}

\clearpage
\section{Additional Attack Samples and Boundary Cases}
\label{app:attack_sample_details}

The previous gallery focuses on successful semantic shifts. This section adds
two complementary views: (i) more paired clean/adversarial examples and (ii)
boundary cases where the attack is weaker. The latter is important because the
perturbation is universal: it cannot adapt to a specific image, so scenes with
strong geometric anchors are more resistant.

\begin{table}[H]
  \caption{Additional attack sample group 1: boundary-style cases with stable
  or weakly changed top-1 labels.}
  \label{tab:additional_boundary_images_group1}
  \centering
  \scriptsize
  \setlength{\tabcolsep}{2pt}
  \resizebox{\textwidth}{!}{%
  \begin{tabular}{@{}M{1.35cm}C{1.55cm}C{1.55cm}M{2.2cm}M{2.2cm}M{4.2cm}@{}}
    \toprule
    Case & Clean image & Adv. image & Clean top-1 & Adv. top-1 & Reading \\
    \midrule
    0507 RS5M &
    \includegraphics[width=1.45cm]{appendix_case_assets/clip_clean/03_idx0507_RS5M_clean.png} &
    \includegraphics[width=1.45cm]{appendix_case_assets/clip_adv/03_idx0507_RS5M_adv.png} &
    beach (0.502) & beach (0.108) &
    The class remains beach, but confidence drops sharply after perturbation. \\
    \midrule
    0377 NWPU &
    \includegraphics[width=1.45cm]{appendix_case_assets/clip_clean/05_idx0377_NWPU_clean.png} &
    \includegraphics[width=1.45cm]{appendix_case_assets/clip_adv/05_idx0377_NWPU_adv.png} &
    white road (0.140) & white road (0.046) &
    The freeway geometry is stable, so the perturbation mainly suppresses confidence. \\
    \midrule
    0953 SkyScript &
    \includegraphics[width=1.45cm]{appendix_case_assets/clip_clean/12_idx0953_SkyScript_clean.png} &
    \includegraphics[width=1.45cm]{appendix_case_assets/clip_adv/12_idx0953_SkyScript_adv.png} &
    white roofs (0.079) & white roofs (0.074) &
    Repeated roof-like structures remain the dominant visual anchor. \\
    \midrule
    0658 RS5M &
    \includegraphics[width=1.45cm]{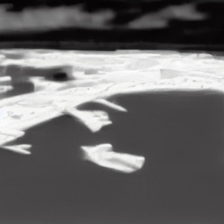} &
    \includegraphics[width=1.45cm]{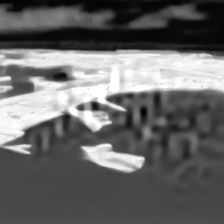} &
    harbor (0.209) & harbor (0.213) &
    Large water and harbor cues remain inside the same semantic neighborhood. \\
    \bottomrule
  \end{tabular}%
  }
\end{table}

\begin{table}[H]
  \caption{Additional attack sample group 2: confidence suppression with nearby
  scene semantics.}
  \label{tab:additional_boundary_images_group2}
  \centering
  \scriptsize
  \setlength{\tabcolsep}{2pt}
  \resizebox{\textwidth}{!}{%
  \begin{tabular}{@{}M{1.35cm}C{1.55cm}C{1.55cm}M{2.2cm}M{2.2cm}M{4.2cm}@{}}
    \toprule
    Case & Clean image & Adv. image & Clean top-1 & Adv. top-1 & Reading \\
    \midrule
    0403 RS5M &
    \includegraphics[width=1.45cm]{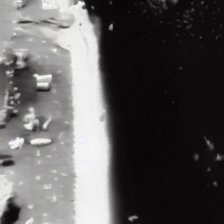} &
    \includegraphics[width=1.45cm]{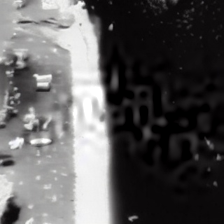} &
    beach (0.414) & beach (0.058) &
    The top-1 remains beach, but the clean confidence is strongly reduced. \\
    \midrule
    0018 RS5M &
    \includegraphics[width=1.45cm]{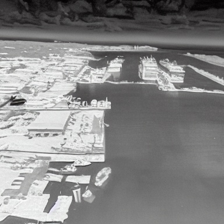} &
    \includegraphics[width=1.45cm]{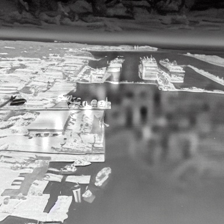} &
    port (0.439) & port (0.146) &
    Port layout survives, but the adversarial input reduces the clean confidence. \\
    \midrule
    0102 RSICD &
    \includegraphics[width=1.45cm]{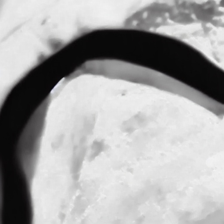} &
    \includegraphics[width=1.45cm]{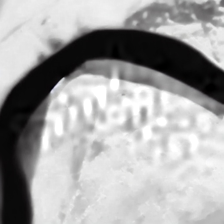} &
    grey river (0.187) & river (0.087) &
    The prediction stays in the same river neighborhood with lower confidence. \\
    \midrule
    0227 RS5M &
    \includegraphics[width=1.45cm]{appendix_case_assets/clip_clean/14_idx0227_RS5M_clean.png} &
    \includegraphics[width=1.45cm]{appendix_case_assets/clip_adv/14_idx0227_RS5M_adv.png} &
    island (0.631) & harbor (0.092) &
    Coastline evidence shifts to a nearby harbor interpretation. \\
    \bottomrule
  \end{tabular}%
  }
\end{table}

\begin{table}[H]
  \caption{Additional attack sample group 3: clear semantic flips.}
  \label{tab:additional_boundary_images_group3}
  \centering
  \scriptsize
  \setlength{\tabcolsep}{2pt}
  \resizebox{\textwidth}{!}{%
  \begin{tabular}{@{}M{1.35cm}C{1.55cm}C{1.55cm}M{2.2cm}M{2.2cm}M{4.2cm}@{}}
    \toprule
    Case & Clean image & Adv. image & Clean top-1 & Adv. top-1 & Reading \\
    \midrule
    0538 RSICD &
    \includegraphics[width=1.45cm]{appendix_case_assets/clip_clean/01_idx0538_RSICD_clean.png} &
    \includegraphics[width=1.45cm]{appendix_case_assets/clip_adv/01_idx0538_RSICD_adv.png} &
    stadium (0.200) & parking lots (0.060) &
    Sports-field evidence is replaced by parking-lot texture. \\
    \midrule
    0516 RSICD &
    \includegraphics[width=1.45cm]{appendix_case_assets/clip_clean/04_idx0516_RSICD_clean.png} &
    \includegraphics[width=1.45cm]{appendix_case_assets/clip_adv/04_idx0516_RSICD_adv.png} &
    rectangular farmland (0.090) & tall buildings (0.186) &
    Smooth field-like evidence turns into building-like response. \\
    \midrule
    0700 RS5M &
    \includegraphics[width=1.45cm]{appendix_case_assets/clip_clean/06_idx0700_RS5M_clean.png} &
    \includegraphics[width=1.45cm]{appendix_case_assets/clip_adv/06_idx0700_RS5M_adv.png} &
    beach (0.253) & tall buildings (0.562) &
    Shoreline evidence is overtaken by high-response structural texture. \\
    \midrule
    0412 RSICD &
    \includegraphics[width=1.45cm]{appendix_case_assets/clip_clean/07_idx0412_RSICD_clean.png} &
    \includegraphics[width=1.45cm]{appendix_case_assets/clip_adv/07_idx0412_RSICD_adv.png} &
    white roofs (0.061) & church (0.207) &
    Roof-like evidence is redirected to an object-like class. \\
    \bottomrule
  \end{tabular}%
  }
\end{table}

\begin{table}[H]
  \caption{Additional attack sample group 4: source-diverse semantic flips.}
  \label{tab:additional_boundary_images_group4}
  \centering
  \scriptsize
  \setlength{\tabcolsep}{2pt}
  \resizebox{\textwidth}{!}{%
  \begin{tabular}{@{}M{1.35cm}C{1.55cm}C{1.55cm}M{2.2cm}M{2.2cm}M{4.2cm}@{}}
    \toprule
    Case & Clean image & Adv. image & Clean top-1 & Adv. top-1 & Reading \\
    \midrule
    0447 NWPU &
    \includegraphics[width=1.45cm]{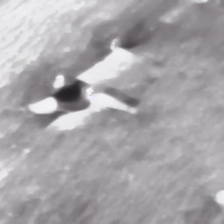} &
    \includegraphics[width=1.45cm]{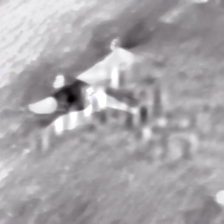} &
    airplane (0.036) & church (0.242) &
    Sparse desert cues are replaced by compact bright structures. \\
    \midrule
    0990 SkyScript &
    \includegraphics[width=1.45cm]{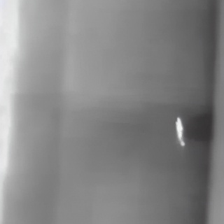} &
    \includegraphics[width=1.45cm]{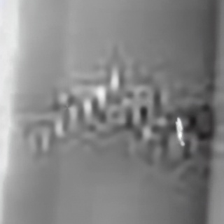} &
    rectangular farmland (0.037) & tall buildings (0.495) &
    Field-like response flips to high-confidence building semantics. \\
    \midrule
    0271 SkyScript &
    \includegraphics[width=1.45cm]{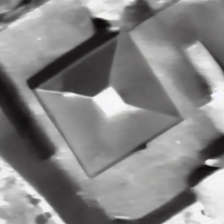} &
    \includegraphics[width=1.45cm]{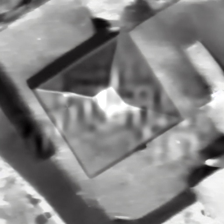} &
    square (0.128) & church (0.379) &
    Compact geometry is reinterpreted as an object-like landmark. \\
    \midrule
    0801 NWPU &
    \includegraphics[width=1.45cm]{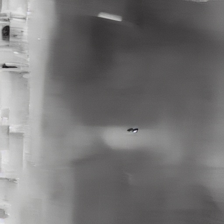} &
    \includegraphics[width=1.45cm]{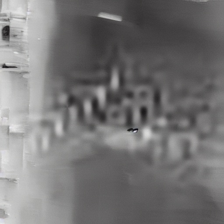} &
    sparse residential (0.035) & tall buildings (0.509) &
    Residential evidence is amplified into tall-building semantics. \\
    \bottomrule
  \end{tabular}%
  }
\end{table}

\begin{table}[H]
  \caption{Representative boundary cases. These classes are not failure modes of
  the implementation; rather, they identify scene types where the universal
  perturbation has less room to override strong geometric evidence.}
  \label{tab:boundary_cases}
  \centering
  \footnotesize
  \resizebox{\textwidth}{!}{%
  \begin{tabular}{@{}cccM{7.2cm}@{}}
    \toprule
    Class or scene type & $n$ & ASR (\%) & Interpretation \\
    \midrule
    Dense residential & 19 & 10.5 &
    Repeated roofs and block layouts provide strong geometric anchors, so the
    airflow texture often changes confidence without flipping the top-1 class. \\
    Storage tanks & 8 & 12.5 &
    Circular tank geometry is visually distinctive and remains stable under the
    smooth perturbation. \\
    Parking lot & 8 & 25.0 &
    Vehicle grids and regular lane structure remain salient in many examples. \\
    Forest & 19 & 31.6 &
    Texture changes can alter confidence, but the global low-contrast vegetation
    pattern often remains inside the same semantic neighborhood. \\
    Tennis court & 8 & 37.5 &
    Court-line geometry is compact and high-contrast, making it harder for a
    universal thermal field to erase completely. \\
    \bottomrule
  \end{tabular}%
  }
\end{table}

\clearpage
\section{Full VLM Prompts and Scoring Protocol}
\label{app:vlm_prompts}

The downstream VLM evaluation uses the same four prompts for all six models and
all attack methods. Captioning is allowed 128 new tokens, while the three
question-answering tasks are allowed 64 new tokens. Generation is deterministic
with greedy decoding (do\_sample=False, num\_beams=1) and each
model's native preprocessing and adapter defaults. The prompts are deliberately
short and task-specific so that changes under attack reflect image-conditioned
model behavior rather than prompt engineering. Table~\ref{tab:vlm_prompts}
lists the exact prompts.

For captioning, we report ROUGE-L against the IR-aware reference caption. The
scoring script also records BLEU, METEOR, and a lightweight CIDEr-style score,
but ROUGE-L is used in the paper because it is stable across the six evaluated
VLMs and directly comparable across methods. For scene recognition, the
generated answer is normalized and matched to the annotated scene label. For
object recognition, the answer is parsed as a set of object or land-cover terms
and compared with the reference object set using F1. For IR-cue evaluation, the
answer is checked against the expected infrared visual cues rather than against a
single class label. The IR-cue prompt is an evaluation prompt designed to elicit
thermal-cue descriptions; the metric should therefore be read as cue-fidelity
under a fixed rubric, not as an unconstrained test of whether a model
independently discovers thermal evidence. We report all clean and adversarial
scores on the same 1{,}000 images and use the statistical protocol described in
Appendix~\ref{app:eval}.

\begin{table}[h]
  \caption{VLM prompts and reported metrics. Lower ROUGE-L, scene accuracy, and
  object F1 indicate stronger degradation; IR-cue accuracy is interpreted
  separately because attacked models may hallucinate plausible thermal cues.}
  \label{tab:vlm_prompts}
  \centering
  \resizebox{\textwidth}{!}{%
  \begin{tabular}{@{}M{2.4cm}M{9.4cm}M{2.0cm}M{2.4cm}@{}}
    \toprule
    Task & Prompt & Max tokens & Metric \\
    \midrule
    Caption &
    Describe this infrared remote sensing image in detail. &
    128 & ROUGE-L \\
    Scene &
    What is the main scene type in this infrared remote sensing image? Answer with a concise scene label only. &
    64 & Scene accuracy \\
    Objects &
    What are the main objects or land-cover elements visible in this infrared remote sensing image? Answer with a concise comma-separated list. &
    64 & Object F1 \\
    IR cues &
    What infrared-specific visual cues can be observed in this image? Mention visible grayscale intensity, high contrast, bright structures, low-texture or dark regions, and structural outlines if present. &
    64 & IR-cue accuracy \\
    \bottomrule
  \end{tabular}%
  }
\end{table}
\clearpage
\section{Per-Source and Per-Class Diagnostic Results}
\label{app:source_diag}

The main paper reports aggregate ASR across models. To test whether the attack
is driven by a single source dataset, we also compute source-wise ASR on the
1{,}000-image diagnostic split. This analysis is secondary because the
diagnostic split is smaller than the full test set, but it is useful for
identifying systematic source effects.

\begin{table}[h]
  \vspace{-1em}
  \caption{Source-wise ASR (\%) on the 1{,}000-image diagnostic split. The same
  images are used for all five CLIP backbones.}
  \label{tab:source_asr}
  \centering
  \resizebox{\textwidth}{!}{%
  \begin{tabular}{@{}cccccccc@{}}
    \toprule
    Source & $n$ & OpenAI-L14 & OpenAI-B32 & OpenCLIP-B32 & RemoteCLIP-B32 & GeoRSCLIP-B32 & Mean \\
    \midrule
    NWPU & 205 & 45.4 & 60.0 & 46.3 & 53.7 & 53.7 & 51.8 \\
    RS5M & 332 & 41.3 & 55.4 & 47.0 & 54.8 & 45.5 & 48.8 \\
    RSICD & 166 & 47.0 & 53.6 & 50.0 & 47.6 & 56.6 & 51.0 \\
    RSITMD & 83 & 48.2 & 57.8 & 59.0 & 50.6 & 59.0 & 54.9 \\
    SkyScript & 214 & 63.6 & 71.0 & 62.1 & 65.0 & 72.9 & 66.9 \\
    \bottomrule
  \end{tabular}%
  }
  \vspace{-1em}
\end{table}

The attack transfers across all five sources, with SkyScript showing the largest
mean ASR and RS5M the smallest. Because the perturbation is universal and not
source-conditioned, this gap most likely reflects differences in image
composition and caption-derived labels rather than a dataset-specific fitting
effect. The source-wise trend is consistent across backbones: all models are
substantially affected on every source, and no source collapses to near-zero
attack success.

We further inspect class-level behavior on the OpenAI-CLIP-B32 diagnostic
classifier output, using only classes with at least five samples after merging
minor spelling variants such as underscore-separated and space-separated labels.
Table \ref{tab:class_extremes} reports representative high- and low-ASR classes.
Classes dominated by broad thermal texture or weak local structure, such as
desert, cloud, beach, playground, and water, are especially vulnerable. Classes
with repeated high-contrast man-made structure, such as dense residential areas,
storage tanks, parking lots, and tennis courts, are more resistant. This
supports the interpretation that AirflowAttack mainly perturbs global thermal
texture and scene-level evidence, while compact geometric structure can still
anchor the model prediction.

\par\medskip
\begingroup
\refstepcounter{table}
\centering
\captionof{table}{Representative high- and low-ASR classes on OpenAI-CLIP-B32 in the
  diagnostic split (classes with $\ge 5$ samples).}
\label{tab:class_extremes}
  \scriptsize
  \setlength{\tabcolsep}{3pt}
  \renewcommand{\arraystretch}{0.95}
  \resizebox{\textwidth}{!}{%
  \begin{tabular}{@{}ll|cccccccc@{}}
    \toprule
    Group & Metric & \multicolumn{8}{c}{Classes} \\
    \midrule
    \multirow{3}{*}{High}
      & Class & desert & cloud & beach & \makecell{play-\\ground} & freeway & school & stadium & mountain \\
      & $n$ & 9 & 6 & 11 & 8 & 7 & 6 & 20 & 13 \\
      & ASR (\%) & 100.0 & 100.0 & 90.9 & 87.5 & 85.7 & 83.3 & 80.0 & 76.9 \\
    \midrule
    \multirow{3}{*}{Low}
      & Class & \makecell{dense\\resid.} & \makecell{storage\\tanks} & palace & \makecell{parking\\lot} & forest & \makecell{medium\\resid.} & \makecell{tennis\\court} & parking \\
      & $n$ & 19 & 8 & 5 & 8 & 19 & 14 & 8 & 28 \\
      & ASR (\%) & 10.5 & 12.5 & 20.0 & 25.0 & 31.6 & 35.7 & 37.5 & 39.3 \\
    \bottomrule
  \end{tabular}%
  }
\par\endgroup
\clearpage

\end{document}